\documentclass{article}
\usepackage{setspace}
\usepackage{indentfirst}
\usepackage{amsmath}
\usepackage{amssymb}
\usepackage{amsthm}
\usepackage{algorithm} %format of the algorithm
\usepackage{algorithmic} %format of the algorithm
\usepackage{multirow} %multirow for format of table
\usepackage{graphicx}
\usepackage{subfigure}
\usepackage{breqn}
\usepackage{multirow} %multirow for format of table
\usepackage{amsfonts}
\usepackage{xcolor}
\usepackage{longtable}
\usepackage{rotating}
\usepackage{booktabs}
\usepackage{threeparttable}
\usepackage{caption}
\usepackage{cases}
\usepackage{lscape}
\usepackage{txfonts}
\usepackage{epstopdf}
\usepackage{authblk}
\usepackage{hyperref}
\usepackage[justification=centering]{caption}
\newtheorem{myDef}{Definition}
\newtheorem{exmp}{Example}%[section]
\newtheorem{rem}{Remark}
\begin{document}

\title{Optimal Randomness in Swarm-based Search}
\author[1]{Jiamin Wei}
\author[1,2]{YangQuan Chen \thanks{Corresponding author: ychen53@ucmerced.edu}}
\author[1]{Yongguang Yu}
\author[3]{Yuquan Chen}
\affil[1]{Department of Mathematics, Beijing Jiaotong University, Beijing 100044, P.R. China}
\affil[2]{School of Engineering, University of California, Merced, 5200 Lake Road, Merced, CA 95343, US}
\affil[3]{Department of Automation, University of Science and Technology of China, Hefei 230027, P.R. China}
\renewcommand*{\Affilfont}{\small\it}
\renewcommand\Authands{ and }
\date{}
\maketitle
\thispagestyle{empty}

\begin{center}
\begin{minipage}{14cm}\vspace{0.1cm}
{\bf Abstract:} L\'{e}vy flights is a random walk where the step-lengths have a probability distribution that is heavy-tailed. It has been shown that L\'{e}vy flights can maximize the efficiency of resource searching in uncertain environments, and also movements of many foragers and wandering animals have been shown to follow a L\'{e}vy distribution. The reason mainly comes from that the L\'{e}vy distribution, has an infinite second moment, and hence is more likely to generate an offspring that is farther away from its parent. However, the investigation into the efficiency of other different heavy-tailed probability distributions in swarm-based searches is still insufficient up to now. For swarm-based search algorithms, randomness plays a significant role in both exploration and exploitation, or diversification and intensification. Therefore, it's necessary to discuss the optimal randomness in swarm-based search algorithms. In this study, CS is taken as a representative method of swarm-based optimization algorithms, and the results can be generalized to other swarm-based search algorithms. In this paper, four different types of commonly used heavy-tailed distributions, including Mittag-Leffler distribution, Pareto distribution, Cauchy distribution, and Weibull distribution, are considered to enhance the searching ability of CS. Then four novel CS algorithms are proposed and experiments are carried out on 20 benchmark functions to compare their searching performances. Finally, the proposed methods are used to system identification to demonstrate the effectiveness.

\vspace{0.2cm}
{\bf Keywords:} optimal randomness; swarm-based search; cuckoo search; heavy-tailed distribution; global optimization
\end{minipage}
\end{center}
\vspace{0.5cm}

\section{Introduction}
Swarm-based search algorithms have attracted great interest of researchers in fields of computational intelligence, artificial intelligence, optimization, data mining, and machine learning during the last two decades \cite{yang2010nature}. Moreover, the swarm intelligence algorithms and artificial intelligence have been successfully used in complex real-life applications, such as wind farm decision system, social aware cognitive radio handovers, feature selection, truck scheduling and so on \cite{anandakumar2018bio,brezovcnik2018swarm,zhao2019research,dulebenets2017novel}. Up to now, a lot of swarm-based search algorithms have been presented, including artificial bee colony (ABC) \cite{karaboga2007powerful}, cuckoo search (CS)\cite{yang2009cuckoo}, firefly algorithm (FA) \cite{yang2009firefly}, particle swarm optimization (PSO) \cite{kennedy1995particle} and so on.

Among the existing swarm-based search algorithms, CS is presented in terms of the obligate brood parasitic behavior of some cuckoo species and the L\'{e}vy flight behavior of some birds and fruit flies. CS searches for new solutions by performing a global explorative random walk together with a local exploitative random walk. CS is famous for utilizing L\'{e}vy flights in its global explorative random walk. L\'{e}vy flights play a critical role in enhancing randomness, as L\'{e}vy flights is a random walk where the step-lengths have a probability distribution that is heavy-tailed. At each iteration process, CS firstly searches for new solutions in L\'{e}vy flights random walk. Secondly, CS proceeds to obtain new solutions in local exploitative random walk. After each random walk, a greedy strategy is used to select a better solution from the current and newly generated solutions according to their fitness values. Due to the salient features such as simple concept, limited parameters, and implementation simplicity, CS has aroused extensive attention and has been accepted as a simple but efficient optimization technique for solving optimization problems. Accordingly, many new CS variants have been continuously presented recently \cite{zheng2012novel,wang2016nearest,rakhshani2017snap,cui2017novel,salgotra2018new}. However, there's still a lot of space in designing newly improved or enhanced techniques to help to increase the accuracy and convergence speed and enhance the searching stability for the original CS algorithm.

In nature, the movements of many foragers and wandering animals have been shown to follow a L\'{e}vy distribution \cite{richer2006levy} rather than Gaussian distribution. It is found that foragers frequently take a large step to enhance the searching efficiency since it is the natural evolution for millions of years. Inspired by the mentioned natural phenomena, CS is proposed by combination with L\'{e}vy, where the step-length is drawn from a heavy-tailed probability distribution and large steps frequently take place flights. In fact, before CS, the idea of L\'{e}vy flights has been applied in \cite{pavlyukevich2007levy} to solve a problem of non-convex stochastic optimization, due to big jumps of the L\'{e}vy flights process. In this way, it can enhance the searching ability compared with Gaussian distribution where large steps seldom happen. More exactly, we have to say the foragers should move following a {\emph{heavy-tailed distribution}} since L\'{e}vy distribution is a simple heavy-tailed distribution which is easy to analyze. There are many other heavy-tailed distributions such as Mittag-Leffler distribution, Pareto distribution, Cauchy distribution, and Weibull distribution, and large steps still frequently happen when using them to generate the steps. For swarm-based optimization algorithms, randomness plays a significant role in both exploration and exploitation, or diversification and intensification \cite{yang2014nature}. Therefore, it's necessary to discuss the optimal randomness in swarm-based search algorithms.

In this paper, we mainly focus on the discussion on the impact of different heavy-tailed distributions on the performance of swarm-based search algorithms. In the study, CS is taken as a representative method of swarm-based optimization algorithms, and the results can be generalized to other swarm-based search algorithms. At first, some basic definitions of the heavy-tailed distributions and how to generate the random numbers according to the given distribution are provided. Then by replacing the L\'{e}vy flight with steps generated from other heavy-tailed distributions, four different randomness-enhanced CS algorithms (namely CSML, CSP, CSC, and CSW) are presented by applying Mittag-Leffler distribution, Pareto distribution, Cauchy distribution and Weibull distribution. Finally, dedicated experimental studies are carried out on a test suite of 20 benchmark problems with unimodal, multimodal, rotated and shifted properties to compare the performance of different variant algorithms. The experimental results demonstrate that the four proposed randomness-enhanced CS algorithms show a significant improvement over the original CS algorithm. This suggests that the performance of CS can be improved by means of integrating different heavy-tailed probability distributions rather than L\'{e}vy flights into it. Moreover, comparisons of CSML, CSP, CSC, and CSW with other optimization algorithms are also performed. At last, an application problem of parameter identification of unknown fractional-order chaotic systems is further considered. Based on the observations and results analysis, the randomness-enhanced CS algorithms are able to exactly identify the unknown specific parameters of the fractional-order system with better effectiveness and robustness. The randomness-enhanced CS algorithms can be regarded as an efficient and promising tool for solving the real-world complex optimization problems besides the benchmark problems.

The remainder of this paper is organized as follows. The principle of the original CS algorithm is described in Section~\ref{CSA}. Section~\ref{RECSA} gives details of four randomness-enhanced CS algorithms after a brief review of several commonly used heavy-tailed distributions. Experimental results and discussions of randomness-enhanced CS algorithms are presented in Section~\ref{ER}. Finally, Section~\ref{CF} summarizes the conclusions and future work.

\section{Cuckoo Search Algorithm}\label{CSA}
Cuckoo search (CS), developed by Yang and Deb, is considered to be a simple but promising stochastic nature-inspired swarm-based search algorithm \cite{yang2009cuckoo,yang2010engineering}. CS is inspired by the intriguing brood parasitism behaviors of some species of cuckoos, and is enhanced by L\'{e}vy flights instead of simple isotropic random walks. Cuckoos are considered to be fascinating birds not only for their beautiful sounds but also for their aggressive reproduction strategy. Some cuckoo species lay their eggs in host nests, and at the same time, they may remove host birds' eggs in order to increase the hatching probability of their own eggs. For simplicity in describing the standard CS, there are three idealized rules as follows \cite{yang2009cuckoo}: (1) Only one egg is laid by each cuckoo bird at a time, and dumped in a randomly chosen nest; (2) The next generations of cuckoos search for new solutions using the best nests with high-quality; (3) The number of available host nests is fixed, and the egg laid by a cuckoo is discovered by the host bird with a probability $P_{a}\in[0,1]$. In this condition, the host bird can either remove the egg or simply abandon the nest and build a completely new nest.

The purpose of CS is to substitute a not-so-good solution in the nests with the new and potentially better solutions (cuckoos). At each iteration process, CS employs a balanced combination of a local random walk and the global explorative random walk under control of a switching parameter $P_{a}$. A greedy strategy is used after each random walk to select better solutions from the current and newly generated solutions based on their fitness values.

%%%%%%%%%%%%%%%%%%%%%%%%%%%%%%%%%%%%%%%%%%
\subsection{L\'{e}vy Flights Random Walk}

%The foraging pattern of the cuckoos is governed by an important factor known as L\'{e}vy flights \cite{brown2007levy}. L\'{e}vy flight models a random walk for large steps, where the step-length is drawn from a heavy-tailed probability distribution. Besides, CS with L\'{e}vy flights based structured random walk has been demonstrated to perform more effective than many existing metaheuristics such as PSO, ABC, and DE \cite{civicioglu2013conceptual}.

At generation $t$, a global explorative random walk carried out by using L\'{e}vy flights can be defined as follows:
\begin{eqnarray}
U_{i}^{t}=X_{i}^{t}+\alpha\otimes {\rm L\acute{e}vy}\otimes(X_{i}^{t}-X_{best}),\label{LFRW}
\end{eqnarray}
where $U_{i}^{t}$ denotes a new solution generated in L\'{e}vy flights random walk, and $X_{best}$ is the best solution obtained so far. $\alpha>0$ is the step size related to the scales of the problem of interest, $X_{best}$ is the best solution obtained so far, the product $\otimes$ represents entrywise multiplications, and ${\rm L\acute{e}vy}(\lambda)$ is defined according to a simple power-law formula as follows:
\begin{eqnarray}
{\rm L\acute{e}vy}(\lambda)\sim t^{-1-\lambda},\label{Levy}
\end{eqnarray}
where $t$ is a random variable, $0<\lambda\leq2$ is a stability index. Moreover, it is worth mentioning that the well-known Gaussian and Cauchy distribution are its special cases when its stability index $\lambda$ is respectively set to 2 and 1.

In practice, ${\rm L\acute{e}vy}(\lambda)$ can be updated as follows:
\begin{eqnarray}
{\rm L\acute{e}vy}(\lambda)\sim \frac{\phi\times\mu}{|v|^{1/\lambda}},
\end{eqnarray}
where $\lambda$ is suggested as 1.5 \cite{yang2010engineering}, $\mu$ and $v$ are random numbers drawn from a normal distribution with mean of 0 and standard deviation of 1, $\Gamma(\cdot)$ denotes the gamma function, and $\phi$ is expressed as:
\begin{eqnarray}
\phi=\left[\frac{\Gamma(1+\lambda)\times \sin(\frac{\pi\times\lambda}{2})}{\Gamma(\frac{1+\lambda}{2})\times\lambda\times2^{\frac{\lambda-1}{2}}}\right]^{1/\lambda}.
\end{eqnarray}

%%%%%%%%%%%%%%%%%%%%%%%%%%%%%%%%%%%%%%%%%%%%%%%%%%%%%%%%%%%%%%%%%%%%%%
%%%%%%%%%%%%%%%% begin figure %%%%%%%%%%%%%%%%%%%
\begin{algorithm}[H]
%\\renewcommand\arraystretch{0.9}%%%%%%%%%%1.0
%\begin{footnotesize}
\caption{\ Pseudo code of the standard CS algorithm}
\label{alg1}
%\setstretch{1.35}
\begin{algorithmic}[1]
%\REQUIRE
%\STATE Initialize parameters: population size $NP$, fraction probability $P_{a}$, dimensionality $D$, the maximum number of function evaluations Max$\_$FEs;
\STATE $t=1$;
\STATE Generate an initial population of $NP$ host nests $X^{t}_{i}$, $(i=1,2,\ldots,NP)$;
\STATE Evaluate the fitness value of each nest $X^{t}_{i}$;
\STATE $FES=NP$;
% \STATE Generate an initial population $X_{i}^{G}=(X_{i,1}^{G},X_{i,2}^{G},\ldots,X_{i,D}^{G})$, $(i=1,2,\ldots,NP)$;
%\STATE Evaluate the fitness value of each nest $X_{i}$;
% \STATE
\STATE Determine the best nest with the best fitness value;
\ \WHILE{FES$<$Max$\_$FEs}
\STATE // \ \textbf{L\'{e}vy flights random walk}
\FOR {$i=1,2,...,NP$}
\STATE Generate a new solution $U_{i}^{t}$ randomly using L\'{e}vy flights random walk according to Equation.~(\ref{LFRW});
%\STATE Search for a new solution $U_{i}^{t}$ using the modified global random walk according to Eqn.~(\ref{MLLFRW});
%\STATE Perform the boundary-handling method;
\STATE Greedily select a better solution from $U_{i}^{t}$ and $X_{i}^{t}$ according to their fitness values;
\STATE $FES=FES+1$;
\ENDFOR
\STATE // \ \textbf{Local random walk, a fraction ($P_{a}$) of worse nests are abandoned and new ones are built}
\FOR {$i=1,2,...,NP$}
\STATE Search for a new solution $U_{i}^{t}$ using local random walk according to Equation.~(\ref{LoRW});
%\STATE Perform the boundary-handling method;
\STATE Greedily select a better solution from $U_{i}^{t}$ and $X_{i}^{t}$ according to their fitness values;
\STATE $FES=FES+1$;
\ENDFOR
%\STATE Keep the best nest with quality solution;
%\STATE Rank the nests and find the current best one;
%\STATE // \ \textbf{Pass the current best nest to the next generation}
\STATE Obtain the best solution so far $X_{best}$;
\STATE $t=t+1$;
\ \ENDWHILE
\STATE Output the best solution.
% \ENSURE the best solution.
\end{algorithmic}
%\end{footnotesize}
\end{algorithm}
%%%%%%%%%%%%%%%% end figure %%%%%%%%%%%%%%%%%%%
%%%%%%%%%%%%%%%%%%%%%%%%%%%%%%%%%%%%%%%%%%%%%%%%%%%%%%%%%%%%%%%%%%%%%%

%%%%%%%%%%%%%%%%%%%%%%%%%%%%%%%%%%%%%%%%%%
\subsection{Local Random Walk}

The local random walk can be defined as:
\begin{eqnarray}
U^{t}_{i}= X^{t}_{i}+r\otimes H(P_{a}-\epsilon)\otimes(X^{t}_{j}-X^{t}_{k}),\label{LoRW}
\end{eqnarray}
where $X^{t}_{j}$ and $X^{t}_{k}$ are two different selected random solutions, $r$ and $\epsilon$ are two independent random numbers with uniform distribution, and $H(u)$ is a Heaviside function. The local random walk utilizes a far field randomization to generate a substantial fraction of new solutions which are sufficiently far from the current best solution. The pseudo-code of the standard CS algorithm is given in Algorithm~\ref{alg1}.

%%%%%%%%%%%%%%%%%%%%%%%%%%%%%%%%%%%%%%%%%%%%%%%%%%%%%%%%%%%%%%%%%%%%%%
\section{Randomness-Enhanced CS Algorithms}\label{RECSA}

The standard CS algorithm uses L\'{e}vy flights in global random walk to explore the search space. The L\'{e}vy step is taken from the L\'{e}vy distribution which is a heavy-tailed probability distribution. In this case, a fraction of large steps is generated, which plays an important role in enhancing the search capability of CS. Although many foragers and wandering animals have been shown to follow a L\'{e}vy distribution \cite{richer2006levy}, the investigation into the impact of other different heavy-tailed probability distributions on CS is still insufficient up to now. This motivates us to make an attempt to apply the well-known Mittag-Leffler distribution, Pareto distribution, Cauchy distribution and Weibull distribution to the standard CS algorithm, by using which, more efficient searches are supposed to take place in the search space thanks to the long jumps. In this section, a brief review of several commonly used heavy-tailed distributions is given, and then the scheme of the randomness-enhanced CS algorithms is introduced.

%%%%%%%%%%%%%%%%%%%%%%%%%%%%%%%%%%%%%%%%%%%%%%%%%%%%%%%%%%%%%%%%%%%%%%
\subsection{Commonly Used Heavy-Tailed Distributions}

This subsection provides the definition of heavy-tailed distribution and several examples of commonly used heavy-tailed distributions.

\begin{myDef}[Heavy-Tailed Distribution]
The distribution of a real-valued random variable $X$ is said to have a heavy right tail if the tail probabilities $P(X>x)$ decay more slowly than those of any exponential distribution, i.e., if
\begin{equation}
\lim_{x\rightarrow\infty}\frac{P(X>x)}{e^{-\lambda x}}=\infty
\end{equation}
for every $\lambda>0$. Heavy left tails are defined in a similar way \cite{foss2011introduction}.
\end{myDef}

\begin{exmp}[Mittag-Leffler Distribution]
A random variable is said to subject to Mittag-Leffler distribution if its distribution function has the following form
%The statistical distribution in terms of the Mittag-Leffler function $E_{\beta}(x)$ was introduced by Pillai \cite{pillai1990mittag}, which are derived according to the distribution function or cumulative density function. The Mittag-Leffler distribution function is defined by
\begin{eqnarray}
F_{\beta}(x)=\sum^{\infty}_{k=1}\frac{(-1)^{k-1}x^{k\beta}}{\Gamma(1+k\beta)},
\end{eqnarray}
where $0<\beta\leq1$, $x>0$, and $F_{\beta}(x)=0$ for $x\leq0$. For $0<\beta<1$, the Mittag-Leffler distribution is a heavy-tailed generalization of the exponential, and reduces to the exponential distribution when $\beta=1$.

A Mittag-Leffler random number can be generated using the most convenient expression proposed by Kozubowski and Rachev \cite{kozubowski1999univariate}:
\begin{eqnarray} \label{MLRN}
\tau_{\beta}=-\gamma\ln u(\frac{\sin(\beta\pi)}{\tan(\beta\pi v)}-\cos(\beta\pi))^{1/\beta},
\end{eqnarray}
where $\gamma$ is the scale parameter, $u,v\in(0,1)$ are independent uniform random numbers, and $\tau_{\beta}$ is a Mittag-Leffler random number.
\end{exmp}

\begin{exmp}[Pareto Distribution]
A random variable is said to subject to Pareto distribution if its cumulative distribution function has the following expression:
\begin{eqnarray}\label{Paretodistribution}
F(x)=\begin{cases}
1-\left(\frac{b}{x}\right)^{a},& \text{$x\geq b$,}\\
0,                         & \text{$x<b$,}
\end{cases}
\end{eqnarray}
where $b>0$ is the scale parameter, $a>0$ is the shape parameter (Pareto's index of inequality).
\end{exmp}

\begin{exmp}[Cauchy Distribution]
A random variable is said to subject to Cauchy distribution if its cumulative distribution function has the following expression:
\begin{eqnarray} \label{Cauchydistribution}
F(x)=\frac{1}{\pi}\arctan\left( \frac{2(x-\mu)}{\sigma} \right)+\frac{1}{2},
\end{eqnarray}
where $\mu$ is the location parameter, $\sigma$ is the scale parameter.
\end{exmp}

\begin{exmp}[Weibull Distribution]
A random variable is said to subject to Weibull distribution if it has a tail function $F$ as follows:
\begin{eqnarray} \label{Weibulldistribution}
F(x)=e^{-(x/\kappa)^{\xi}},
\end{eqnarray}
where $\kappa>0$ is the scale parameter, $\xi>0$ is the shape parameter. If and only if $\xi<1$, the Weibull distribution is a heavy-tailed distribution.
\end{exmp}

%%%%%%%%%%%%%%%%%%%%%%%%%%%%%%%%%%%%%%%%%%%%%%%%%%%%%%%%%%%%%%%%%%%%%%
\subsection{Improving CS with Different Heavy-Tailed Probability Distributions} \label{CSHT}
For swarm-based search algorithms, randomness plays a significant role in both exploration and exploitation, or diversification and intensification \cite{yang2014nature}. It's very necessary to discuss the optimal randomness in swarm-based search algorithms. Randomness is normally realized by employing pseudorandom numbers, based on some common stochastic processes. Generally, randomization is achieved by simple random numbers that are drawn from a uniform distribution or a normal distribution. But in other cases, more sophisticated randomization approaches are considered, for example, random walks and L\'{e}vy flights. Here, we have to say the foragers should move following a {\emph{heavy-tailed distribution}} since L\'{e}vy distribution is a simple heavy-tailed distribution which is easy to analyze. There are many other heavy-tailed distributions such as Mittag-Leffler distribution, Pareto distribution, Cauchy distribution, and Weibull distribution, and large steps still frequently happen when using them to generate the steps. In this paper, we mainly focus on the discussion on the impact of different heavy-tailed distributions on the performance of swarm-based search algorithms. In the study, CS is taken as a representative method of swarm-based optimization algorithms, and the results can be generalized to other swarm-based search algorithms.

In this section, four randomness-enhanced cuckoo search algorithms are proposed in this paper. Specifically, the following modified CS methods are considered: (1) CS with the Mittag-Leffler distribution, denoted as CSML; (2) CS with the Pareto distribution, denoted as CSP; (3) CS with the Cauchy distribution, denoted as CSC; (4) CS with the Weibull distribution, referred to CSW. In the modified CS methods, the aforementioned four different heavy-tailed probability distributions are respectively used to be integrated into CS instead of the original L\'{e}vy flights in the global explorative random walk. By using these heavy-tailed probability distributions, the updating equation~(\ref{LFRW}) can be reformulated as follows
\begin{eqnarray} \label{MLLFRW}
U_{i}^{t}=X_{i}^{t}+\alpha\otimes {\rm Mittag-Leffler}(\beta,\gamma)\otimes(X_{i}^{t}-X_{best}),
\end{eqnarray}
\begin{eqnarray} \label{ParetoLFRW}
U_{i}^{t}=X_{i}^{t}+\alpha\otimes {\rm Pareto}(b,a)\otimes(X_{i}^{t}-X_{best}),
\end{eqnarray}
\begin{eqnarray} \label{CauchyLFRW}
U_{i}^{t}=X_{i}^{t}+\alpha\otimes {\rm Cauchy}(\mu,\sigma)\otimes(X_{i}^{t}-X_{best}),
\end{eqnarray}
\begin{eqnarray} \label{WeibullLFRW}
U_{i}^{t}=X_{i}^{t}+\alpha\otimes {\rm Weibull}(\xi,\kappa)\otimes(X_{i}^{t}-X_{best}),
\end{eqnarray}
where ${\rm Mittag-Leffler}(\beta,\gamma)$ in Equation~(\ref{MLLFRW}) denotes a random number drawn from Mittag-Leffler distribution; ${\rm Pareto}(b,a)$ in Equation~(\ref{ParetoLFRW}) represents a random number drawn from Cauchy distribution; ${\rm Cauchy}(\mu,\sigma)$ in Equation~(\ref{CauchyLFRW}) denotes a random number drawn from Cauchy distribution; ${\rm Weibull}(\alpha,\kappa)$ in Equation~(\ref{WeibullLFRW}) means a random number drawn from Weibull distribution. Compared with the standard CS algorithm, the differences of randomness-enhanced cuckoo search methods lie in line 9 from Algorithm 1. % New solutions are generated using different heavy-tailed probability distributions according to the Equations~(\ref{MLLFRW})~(\ref{ParetoLFRW})~(\ref{CauchyLFRW})~(\ref{WeibullLFRW}).

\begin{rem}
In this paper, our emphasis is to study the effects of different heavy-tailed distributions on the swarm-based search algorithms.
\end{rem}

\begin{rem}
Since CS is a popular swarm-based search algorithm, we only use it as an representative. Similar analyses for optimal randomness can be applied to other swarm-based algorithms.
\end{rem}

\begin{rem}
The source code of randomness-enhanced cuckoo search algorithms (namely CSML, CSP, CSC, CSW), written in Matlab, is available at \\ \url{https://www.mathworks.com/matlabcentral/fileexchange/71758-optimal-randomness-in-swarm-based-search}.
\end{rem}

%%%%%%%%%%%%%%%%%%%%%%%%%%%%%%%%%%%%%%%%%%%%%%%%%%%%%%%%%%%%%%%%%%%%%%
\section{Experimental Results}\label{ER}

This study focuses on discussing the effectiveness and efficiency of the proposed randomness-enhanced CS algorithms. To fulfill this purpose, extensive experiments are carried out on a test suite of 20 benchmark functions, which are chosen from the literature \cite{noman2008accelerating,suganthan2005problem}. The superiority of randomness-enhanced CS algorithms over the standard CS is tested, then a scalability study and comparison with other optimization algorithms are performed. Finally, an application to parameter identification of fractional-order chaotic systems is also investigated.

%%%%%%%%%%%%%%%%%%%%%%%%%%%%%%%%%%%%%%%%%%%%%%%%%%%%%%%%%%%%%%%%%%%%%%
\subsection{Experimental Setup}

For parameter settings of CS, CSML, CSP, CSC and CSW, the probability $P_{a}$ is set to 0.25 \cite{yang2009cuckoo}, the scaling factor $\alpha$ is set to 0.01. The proposed randomness-enhanced CS algorithms introduce new parameters to CS: the scale parameter $\gamma$ and the Mittag-Leffler index $\beta$ in CSML; the scale parameter $b$ and the shape parameter $a$ in CSP; the location parameter $\mu$ and the scale parameter $\sigma$ in CSC; the scale parameter $\kappa>0$ and the shape parameter $\xi$ in CSW. As for these newly introduced parameters, their values are given in Table~\ref{Paraforalgs} after analysis in Section~\ref{PT}. Moreover, the population size satisfies $NP=D$ where $D$ denotes the dimension of the problem unless a change is mentioned. In the experimental studies, the maximum number of function evaluations (namely Max$\_$FEs) is taken as the termination criterion and set to $10,000\times D$. All the algorithms are evaluated for 50 times and the averaged experimental results are recorded for each benchmark function respectively. Besides, two non-parametric statistical tests for independent samples are taken to detect the differences between the proposed algorithm and the compared algorithms. The tests contain the Wilcoxon signed-rank test at the 5\% significance level and the Friedman test. The symbol "\ddag", "\dag" and "=" respectively denote the average performance gained by the chosen approach is weaker than, better than, and similar to the compared algorithm. Meanwhile, the best experimental results for each benchmark problem are marked in boldface, for clarity.

%% table 1
\begin{table}[htbp]
\renewcommand\arraystretch{0.8}
\begin{center}
\caption{Parameters for randomness-enhanced CS algorithms.}\label{Paraforalgs}
\begin{tabular*}{\hsize}{@{}@{\extracolsep{\fill}}lllllllll@{}} \toprule
Distribution &Algorithm &Parameters   \\ \midrule
Mittag-Leffler distribution &CSML &$\gamma=4.5$, $\beta=0.8$ \\
Pareto distribution         &CSP  &$a=1.5$, $b=4.5$ \\
Cauchy distribution         &CSC  &$\sigma=4.5$, $\mu=0.8$ \\
Weibull distribution        &CSW  &$\xi=0.3$, $\kappa=4$ \\
\bottomrule
\end{tabular*}
\end{center}
\end{table}

\begin{figure}[htbp]
\centering
\subfigure[CSML]{
\begin{minipage}[t]{0.45\linewidth}
\centering
\includegraphics[width=2.2in]{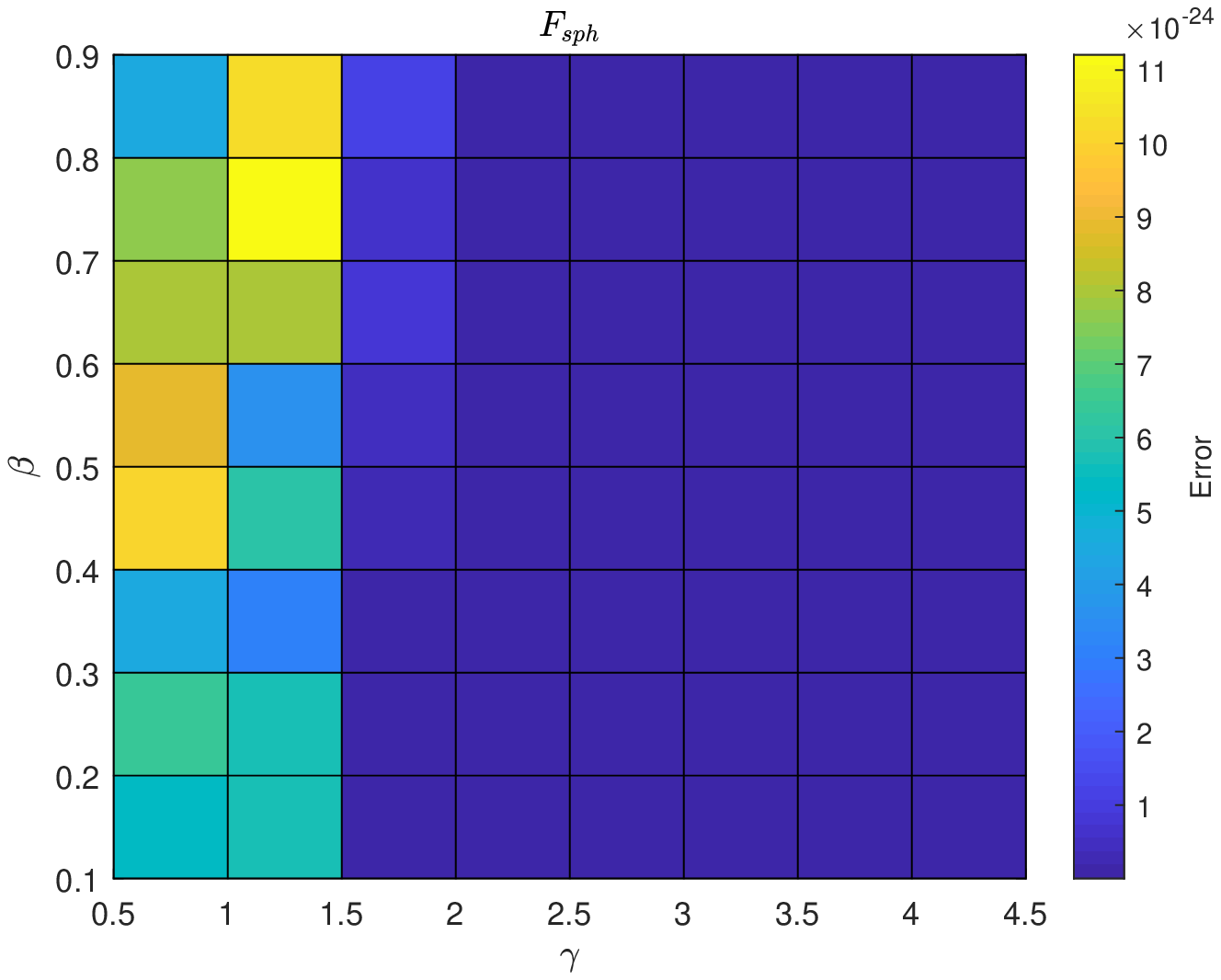}
\end{minipage}%\label{figproML1}
\begin{minipage}[t]{0.45\linewidth}
\centering
\includegraphics[width=2.2in]{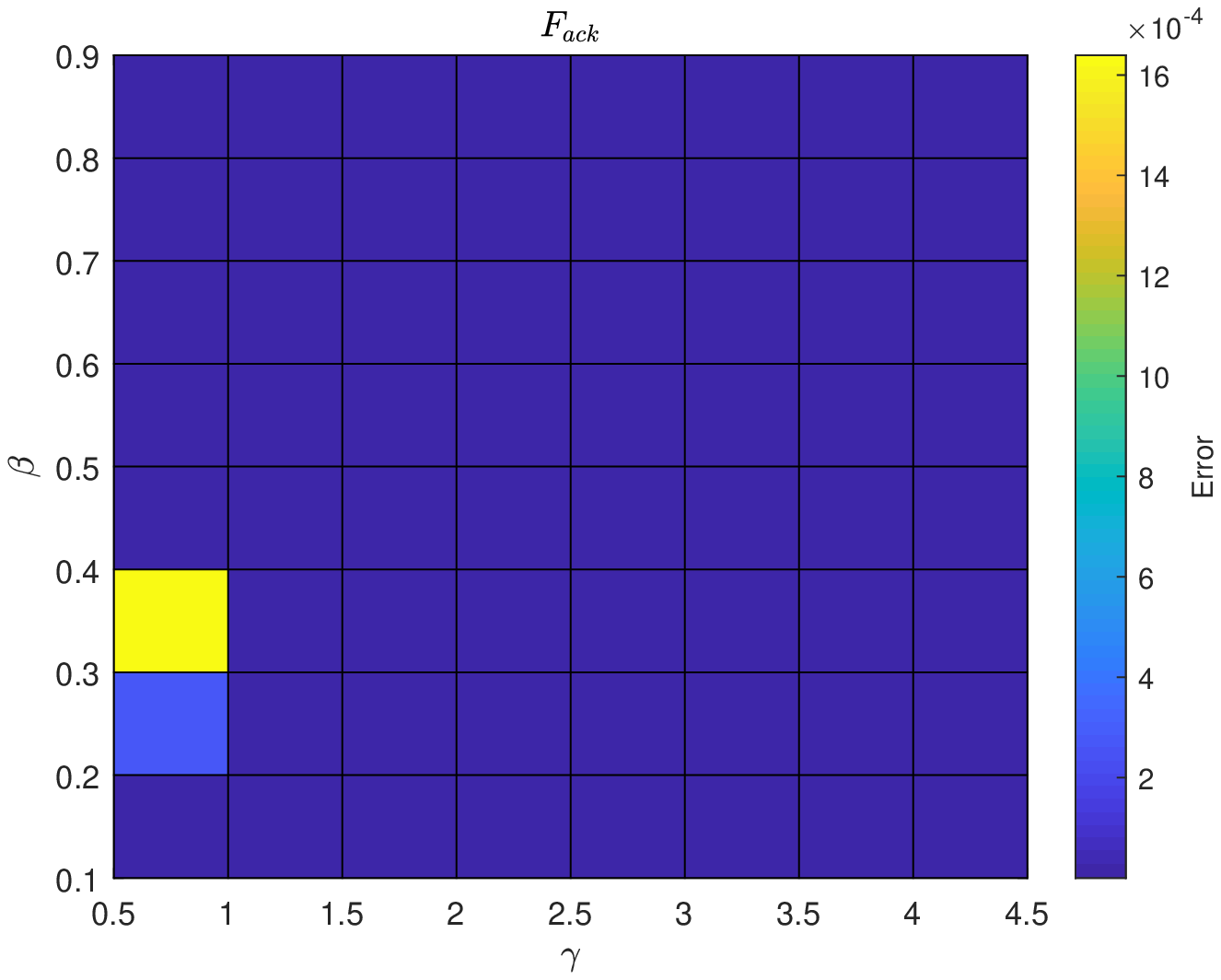}
\end{minipage}\label{figproML}%\label{figproML2}
}
\subfigure[CSP]{
\begin{minipage}[t]{0.45\linewidth}
\centering
\includegraphics[width=2.2in]{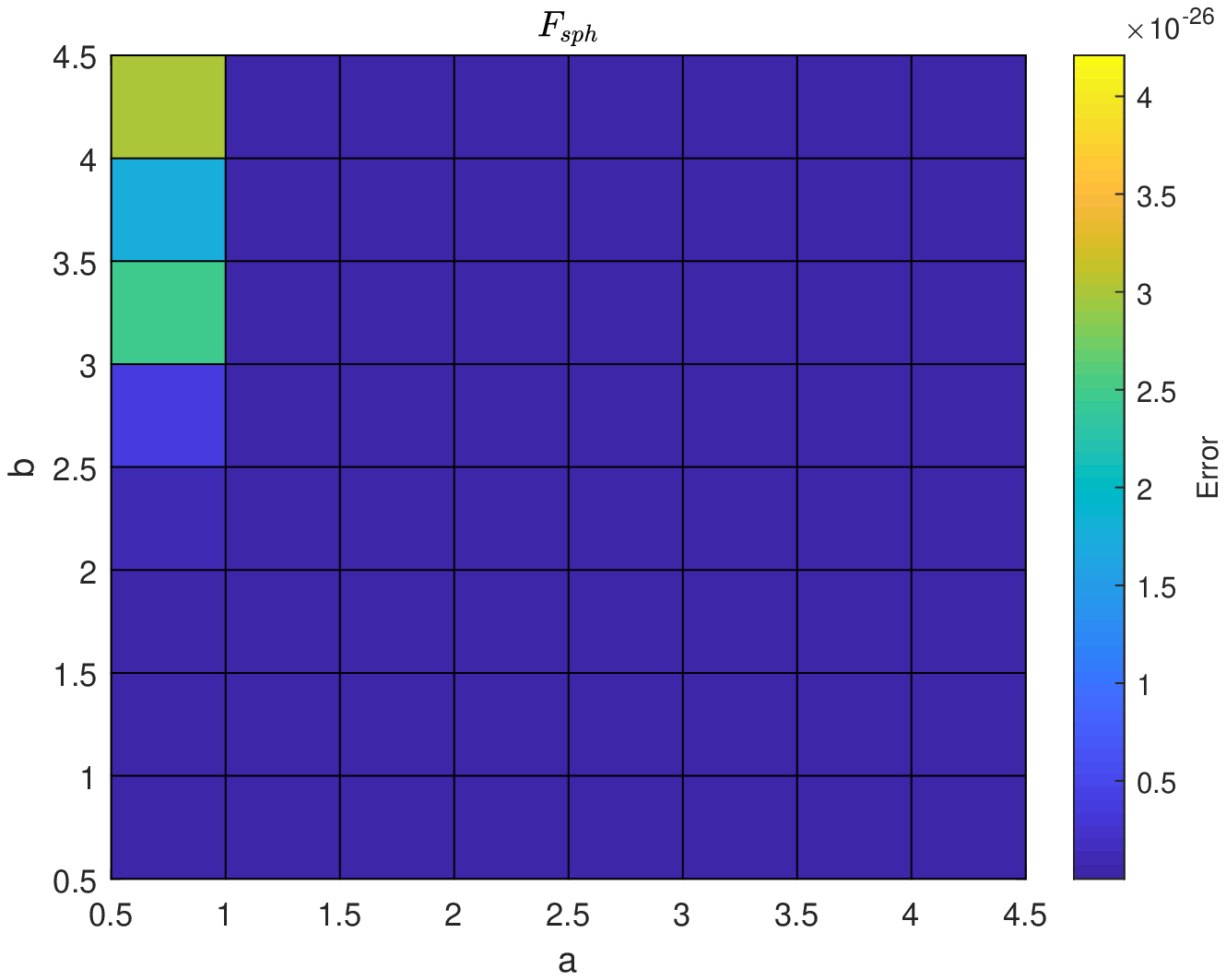}
\end{minipage}%\label{figproP1}
\begin{minipage}[t]{0.45\linewidth}
\centering
\includegraphics[width=2.2in]{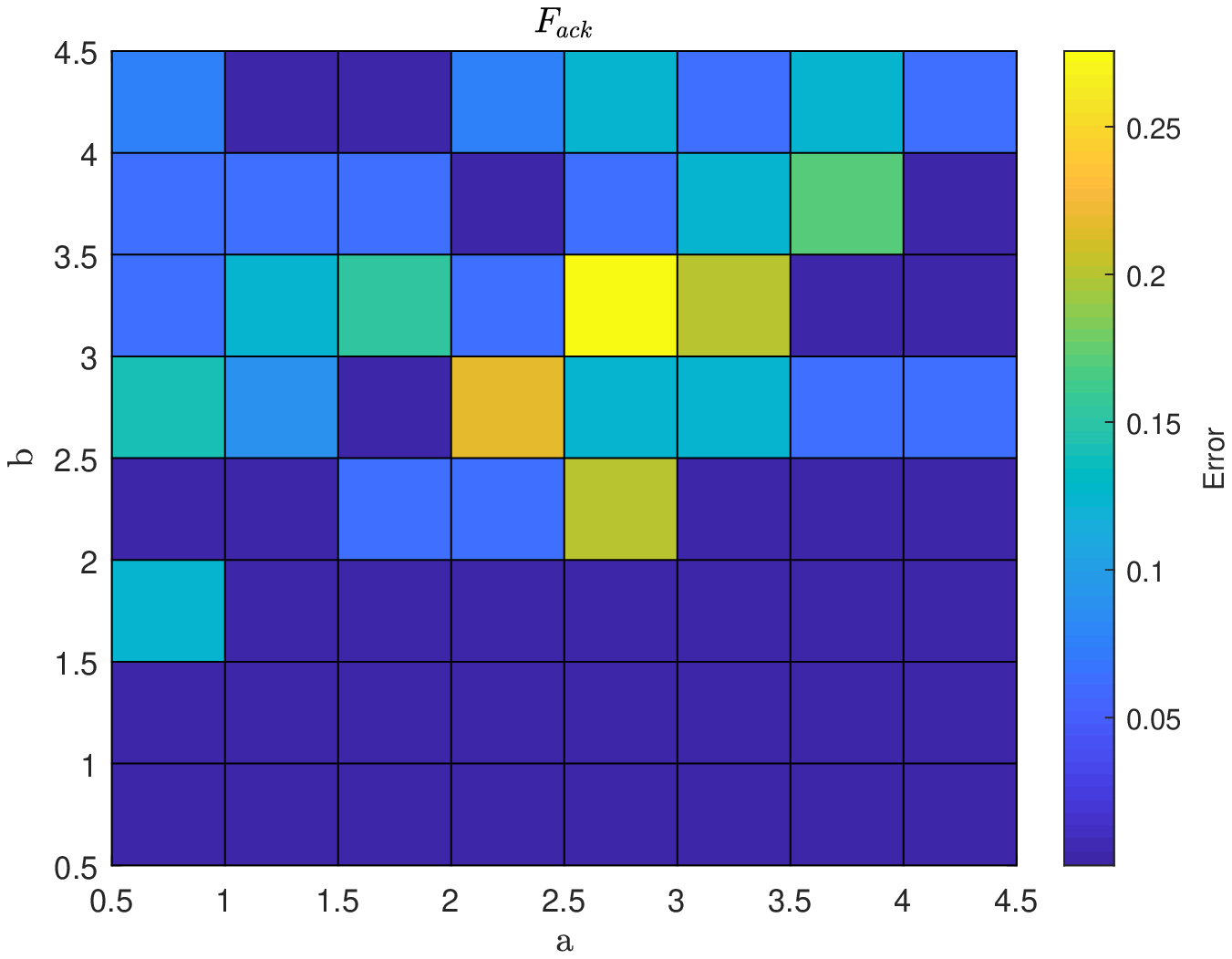}
\end{minipage}\label{figproP}%\label{figproP2}
}
\subfigure[CSC]{
\begin{minipage}[t]{0.45\linewidth}
\centering
\includegraphics[width=2.2in]{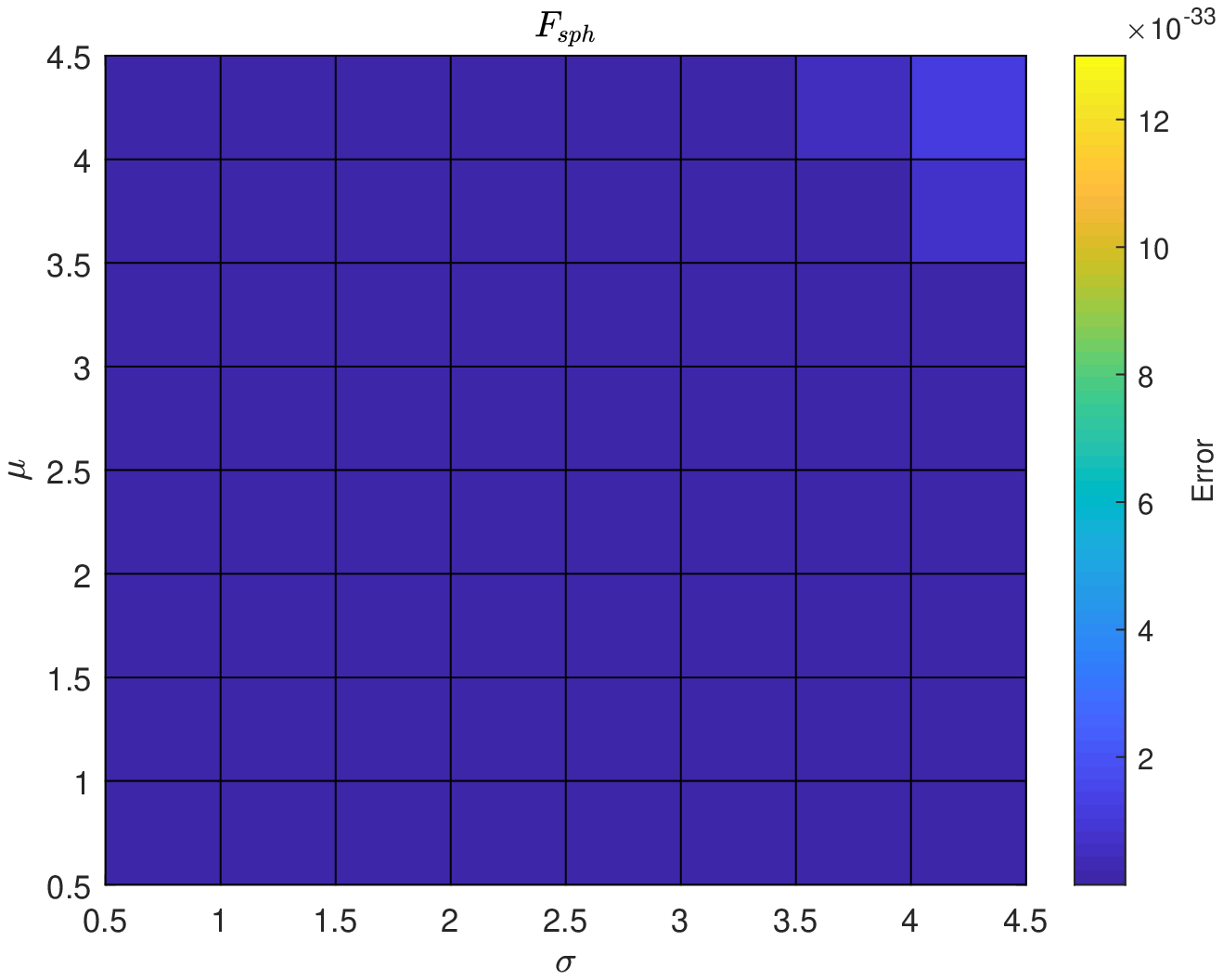}
\end{minipage}%\label{figproC1}
\begin{minipage}[t]{0.45\linewidth}
\centering
\includegraphics[width=2.2in]{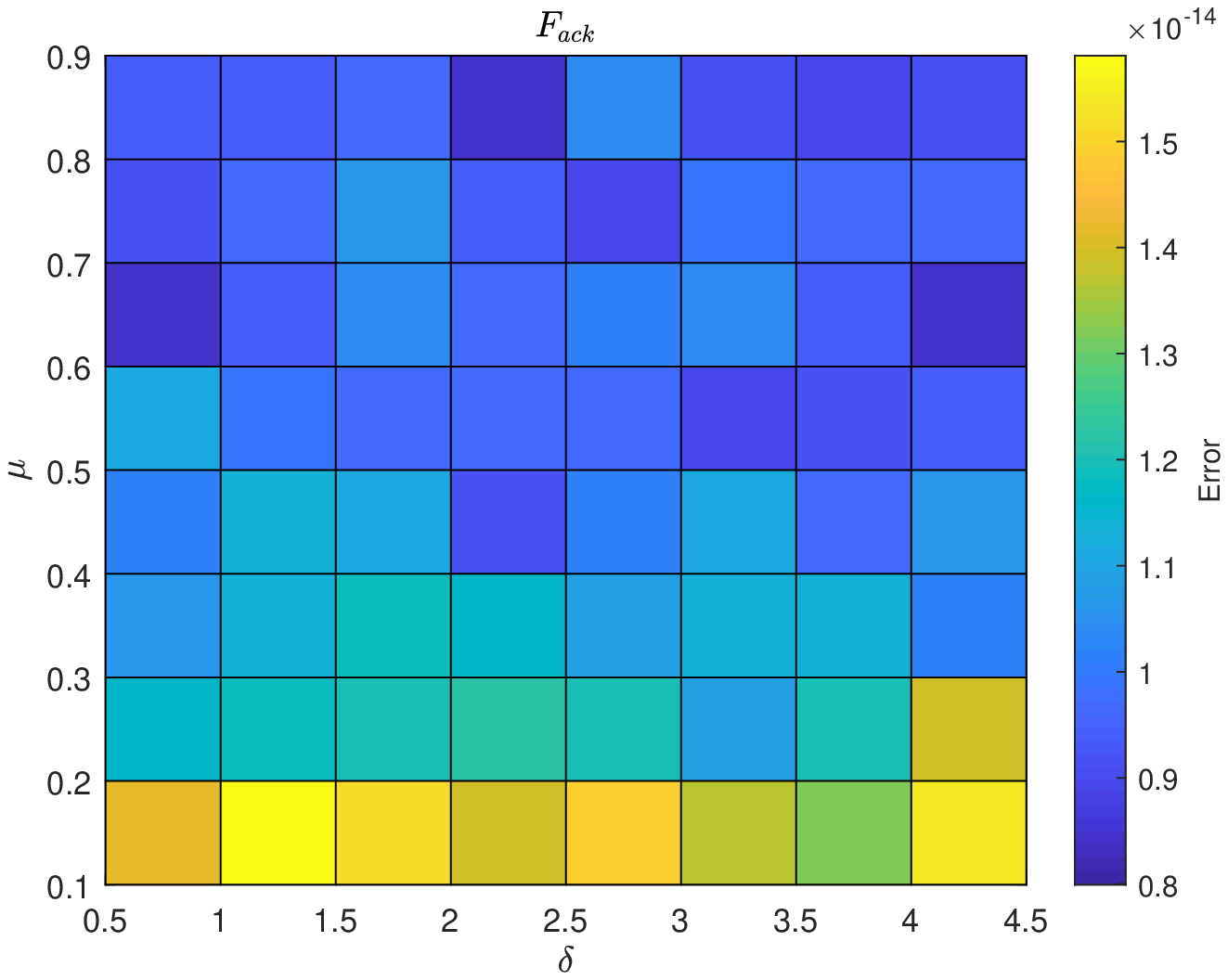}
\end{minipage}\label{figproC}
}
\subfigure[CSW]{
\begin{minipage}[t]{0.45\linewidth}
\centering
\includegraphics[width=2.2in]{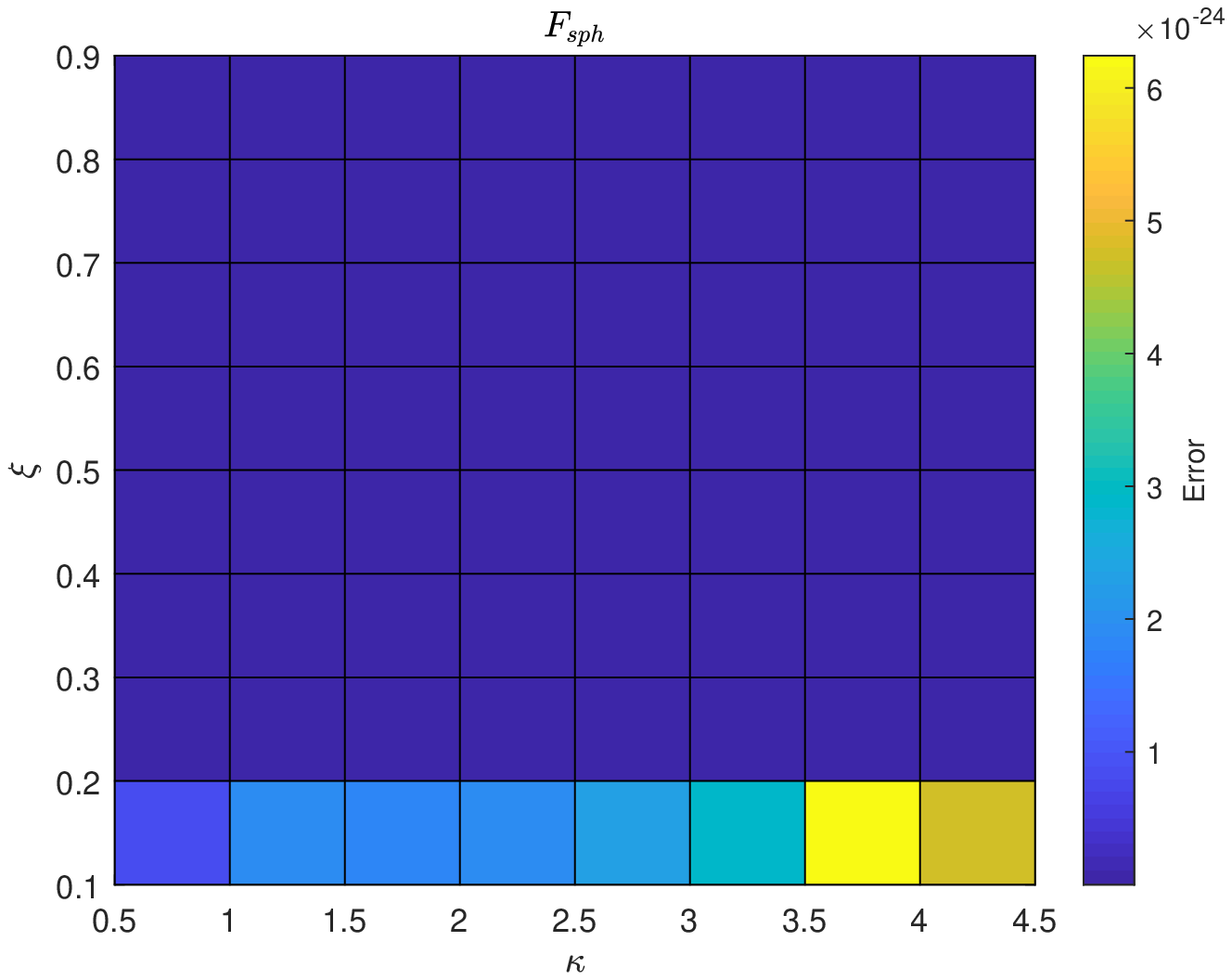}
\end{minipage}%\label{figproW1}
\begin{minipage}[t]{0.45\linewidth}
\centering
\includegraphics[width=2.2in]{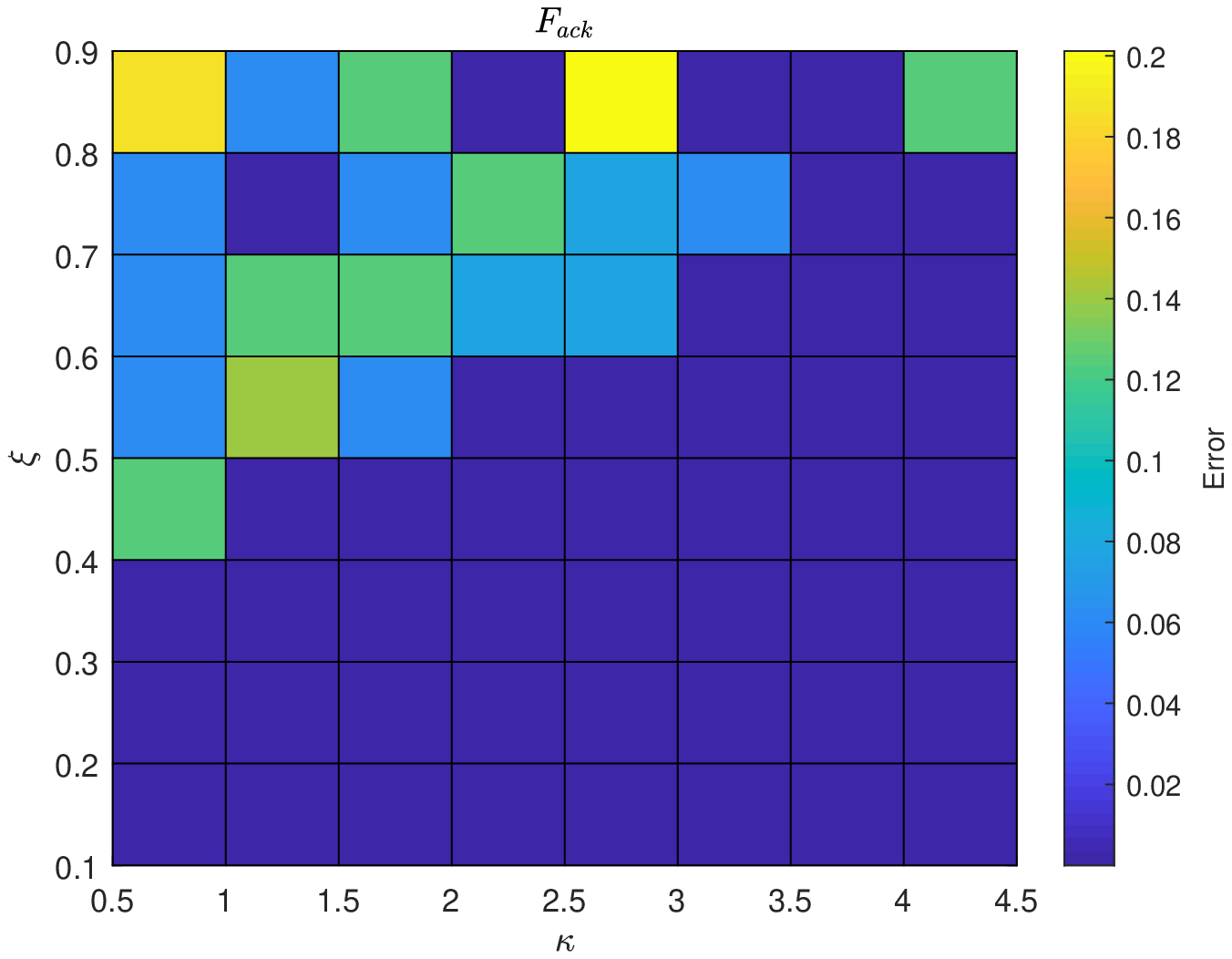}
\end{minipage}\label{figproW}
}
\caption{Impact of user-defined parameter values of CSML, CSP, CSC and CSW on the results for selected benchmark functions.}
\label{figparam}
\end{figure}
%%%%%%%%%%%%%%%% end figure %%%%%%%%%%%%%%%%%%%
%%%%%%%%%%%%%%%%%%%%%%%%%%%%%%%%%%%%%%%%%%%%%%%%%%%%%%%%%%%%%%%%%%%%%%

%%%%%%%%%%%%%%%%%%%%%%%%%%%%%%%%%%%%%%%%%%%%%%%%%%%%%%%%%%%%%%%%%%%%%%
\subsection{Parameter Tuning} \label{PT} %Parameter Analysis
From Section~\ref{CSHT}, it's obvious that each of the four randomness-enhanced CS algorithms brings two new user-defined parameters, for example, the scale parameter $\gamma$ and the Mittag-Leffler index $\beta$ in CSML. To illustrate the impact of these two parameters on the optimization results and to offer reference values to users of our algorithm, parameter analyses are conducted in advance and corresponding experiments are performed on unimodal function $F_{sph}$ and multimodal function $F_{ack}$ with dimension $D$ set to 30. The optimal value of selected benchmark functions is 0. $10,000\times D$ is the default value for Max$\_$FEs. 15 independent runs are carried out for each parameter setting to reduce statistical sampling effects. The experimental results are plotted in Figure~\ref{figparam}. For simplicity of description, only the result of parameter tuning for CSML is shown here, and the same operation is conducted on CSP, CSC, and CSW. In Figure~\ref{figproML}, $\gamma$ varies within interval $[0.5,4.5]$ in steps of 0.5, $\beta$ varies from 0.1 to 0.9 in steps of 0.1, and `Error' represents the average error to the optimal value.

From Figure~\ref{figproML}, we can see that the Mittag-Leffler index $\beta$, in general, has a slight effect on the performance of CSML, whereas the value of scale parameter $\gamma$ shows a more significant impact on the experimental results. According to the right part of each subfigure in Figure~\ref{figproML}, the larger the value of scale parameter $\gamma$ is, the better the performance of CSML will be. In view of the above considerations, we set the values of $\gamma$ and $\beta$ to 0.8 and 4.5 for all the experiments being conducted in the next subsections. For Pareto distribution, Cauchy distribution and Weibull distribution, the same parameter analysis is performed according to Figures~\ref{figproP}~\ref{figproC}~\ref{figproW}. The user-defined parameter values for all the randomness-enhanced CS algorithms are listed in Table~\ref{Paraforalgs}.

%%%%%%%%%%%%%%%%%%%%%%%%%%%%%%%%%%%%%%%%%%%%%%%%%%%%%%%%%%%%%%%%%%%%%%
%\subsection{Effectiveness of Different Heavy-Tailed Probability Distributions on CS}
\subsection{Performance Evaluation of Randomness-Enhanced CS Algorithms} \label{PE}
In this section, lots of experiments are performed in order to probe into the effectiveness and efficiency of different heavy-tailed distributions on the performance of CS, and meanwhile, to decide the optimal randomness in improving CS.
In our experiments, the standard CS and four proposed randomness-enhanced CS algorithms (namely, CSML, CSP, CSC, and CSW) are tested on 20 test functions where $D$ is set to 30. The experimental results are presented in Table~\ref{Compare}.

According to Table~\ref{Compare}, it can be clearly found that CS with different heavy-tailed probability distributions provides significantly better results when compared with the original CS. Specifically speaking, in terms of the total number of "$\ddagger/\approx/\dagger$", CS is inferior to CSML, CSP, CSC, and CSW on 17, 17, 16 and 16 test functions, similar to CSML, CSP, CSC and CSW on 1, 1, 2 and 1 test functions, and superior to CSML, CSP, CSC, and CSW on 2, 2, 2 and 3 test functions, respectively. It is worth noting that CSML, CSP, CSC and CSW are capable of achieving the global optimum on test problem $F_{grw}$ and $F_{1}$, while CS doesn't. Moreover, all the p-values are less than 0.05. These results suggest that CSML, CSP, CSC, and CSW are able to significantly improve the performance of CS for the test functions at $D=30$. The comprehensive ranking orders are CSW, CSC, CSML, CSP, and CS in a descending manner. This indicates that the integration of different heavy-tailed probability distributions into CS not only retains the merit of CS, but also performs even better. Besides, the Weibull distribution performs the best in enhancing the search ability of CS, that is, CSW is supposed to be the optimal randomness in improving CS among all the comparison methods for solving benchmark problems at $D=30$.

%%%%%%%%%%%%%%%%%%%%%%%%%%%%%%%%%%%%%%%%%%%%%%%%%%%%%%%%%%%%%%%%%%%%%%
%%%%%%%%%%%%%%%% begin figure %%%%%%%%%%%%%%%%%%%
\begin{figure}[htbp]
\centering
\subfigure[]{
%\subfigure[pic1.]{
\begin{minipage}[t]{0.45\linewidth}
\centering
\includegraphics[width=2.5in]{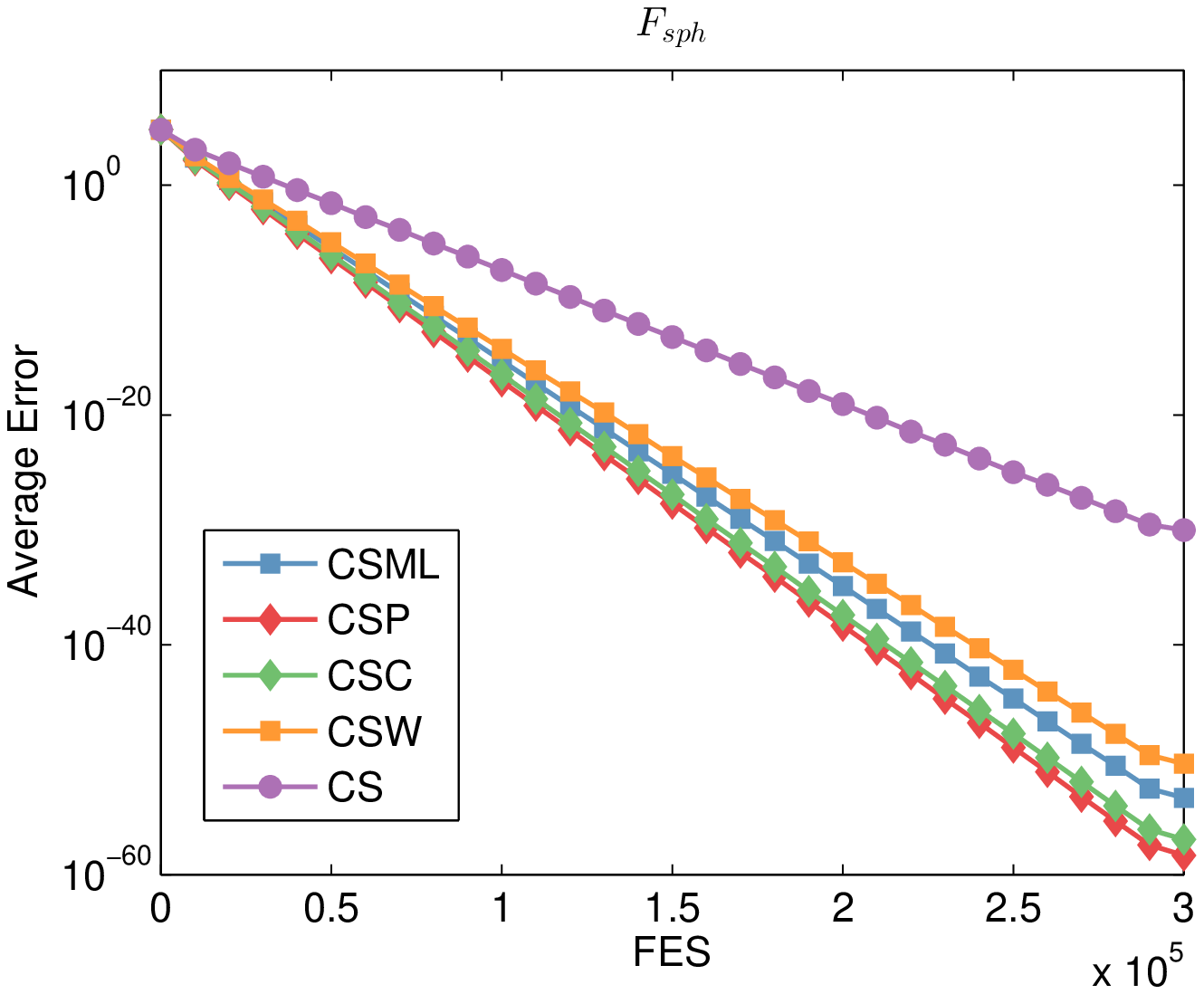}
%\caption{fig1}
\end{minipage}%
}%
\subfigure[]{
%\subfigure[pic2.]{
\begin{minipage}[t]{0.45\linewidth}
\centering
\includegraphics[width=2.5in]{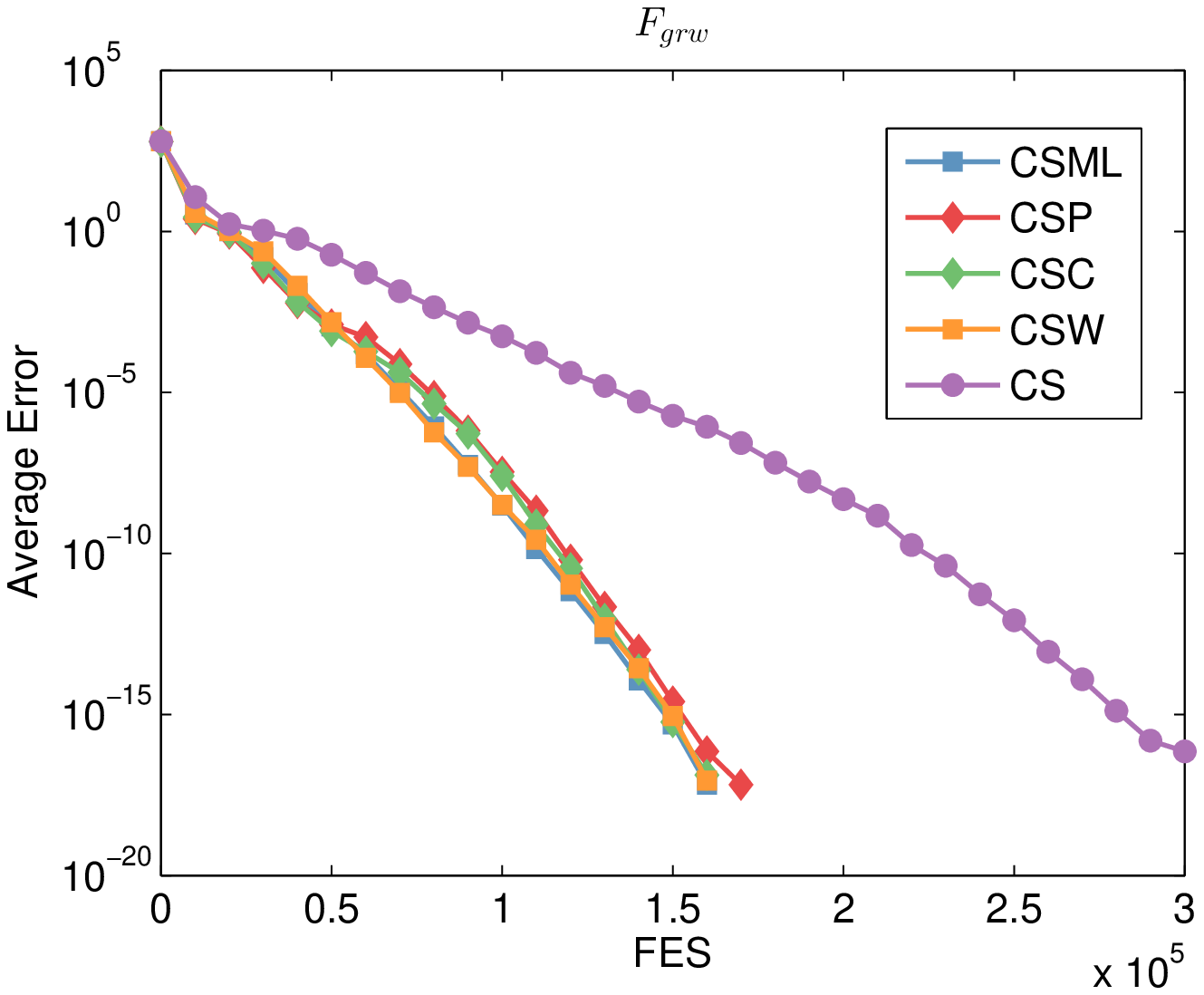}
%\caption{fig2}
\end{minipage}%
}%
\\
\subfigure[]{
\begin{minipage}[t]{0.45\linewidth}
\centering
\includegraphics[width=2.5in]{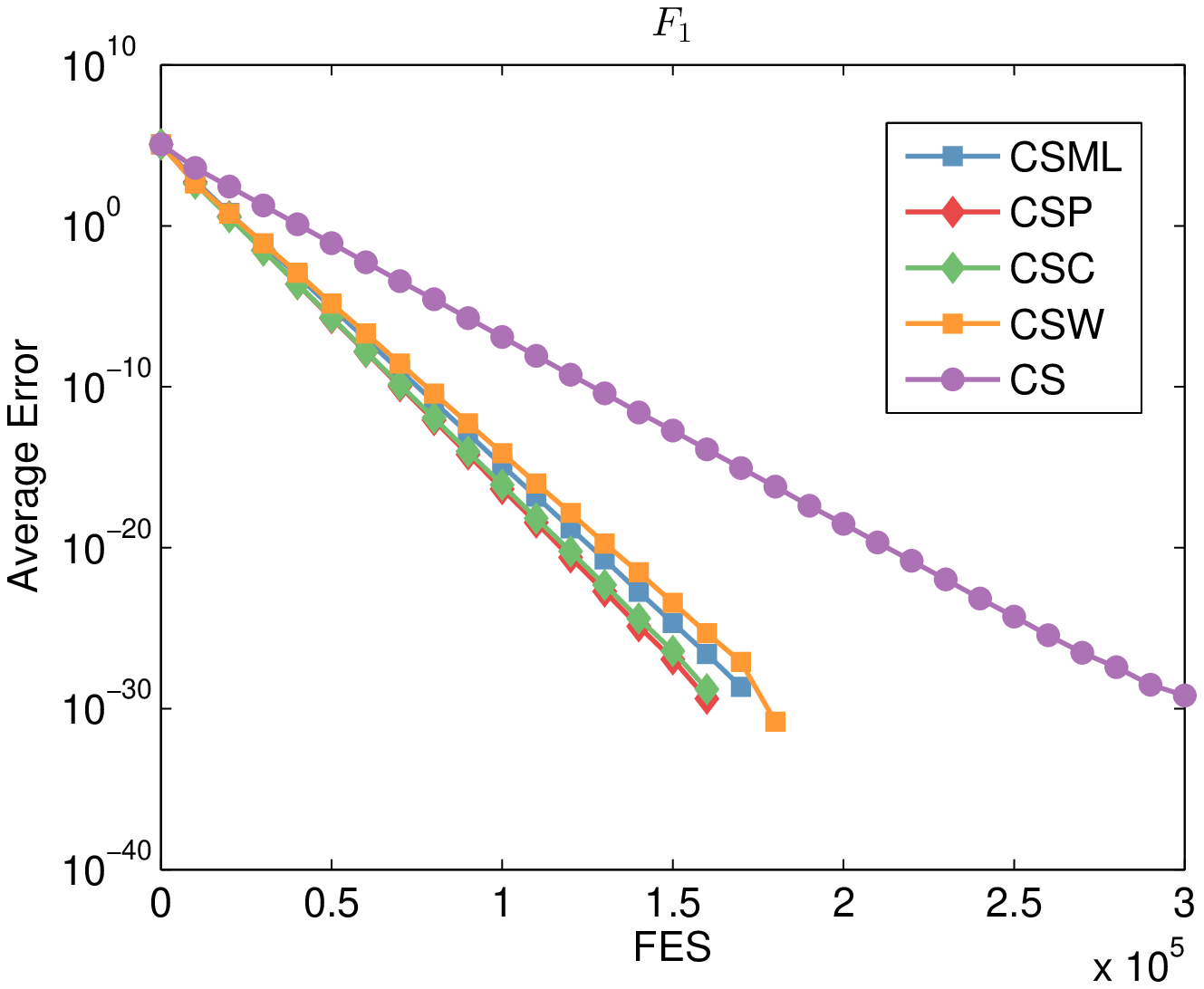}
%\caption{fig1}
\end{minipage}%
}%
\subfigure[]{
%\subfigure[pic2.]{
\begin{minipage}[t]{0.45\linewidth}
\centering
\includegraphics[width=2.5in]{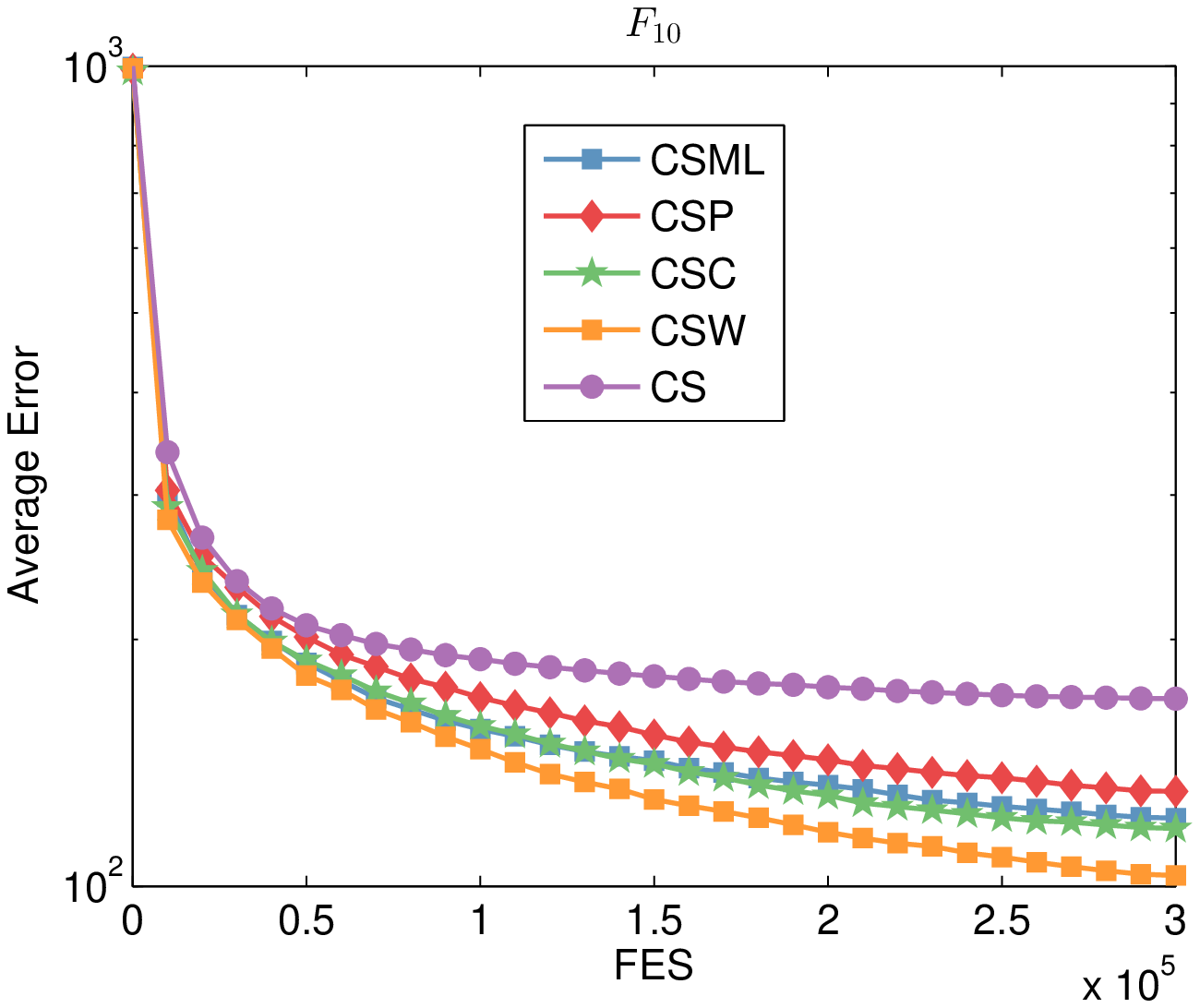}
%\caption{fig2
\end{minipage}%
}%
\caption{Convergence curves of CS and different improved CS algorithms for selected functions at $D=30$.}
\label{figcompare}
\end{figure}
%%%%%%%%%%%%%%%% end figure %%%%%%%%%%%%%%%%%%%
%%%%%%%%%%%%%%%%%%%%%%%%%%%%%%%%%%%%%%%%%%%%%%%%%%%%%%%%%%%%%%%%%%%%%%

\begin{table}[htbp]
\renewcommand\arraystretch{0.8}
\caption{Comparisons between CS and four randomness-enhanced CS algorithms at $D=30$.}\label{Compare}
\begin{center}
%\resizebox{\textwidth}{!}{
\begin{threeparttable}
\begin{tabular*}{\hsize}{@{}@{\extracolsep{\fill}}lllllllll@{}} \toprule
%\begin{tabular}{lllllllll} \toprule
Fun &CS  &CSML &CSP &CSC &CSW  \cr \midrule			
$F_{sph}$	&9.58E-31	&4.90E-54$^{\ddagger}$	&\textbf{4.74E-59}$^{\ddagger}$	&1.17E-57$^{\ddagger}$	&4.40E-51$^{\ddagger}$\\
$F_{ros}$	&1.20E+01	&5.22E+00$^{\ddagger}$	&3.10E+00$^{\ddagger}$	&\textbf{2.74E+00}$^{\ddagger}$	&8.62E+00$^{\ddagger}$\\
$F_{ack}$	&7.70E-13	&1.06E-14$^{\ddagger}$	&1.07E-14$^{\ddagger}$	&9.56E-15$^{\ddagger}$	&\textbf{8.28E-15}$^{\ddagger}$\\
$F_{grw}$	&7.11E-17	&\textbf{0.00E+00}$^{\ddagger}$	&\textbf{0.00E+00}$^{\ddagger}$	&\textbf{0.00E+00}$^{\ddagger}$	&\textbf{0.00E+00}$^{\ddagger}$\\
$F_{ras}$	&2.32E+01	&1.38E+01$^{\ddagger}$	&1.88E+01$^{\ddagger}$	&1.49E+01$^{\ddagger}$	&\textbf{8.34E+00}$^{\ddagger}$\\
$F_{sch}$	&1.57E+03	&5.37E+02$^{\ddagger}$	&1.32E+03$^{\ddagger}$	&4.80E+02$^{\ddagger}$	&\textbf{3.56E+01}$^{\ddagger}$\\
$F_{sal}$	&3.76E-01	&2.96E-01$^{\ddagger}$	&3.00E-01$^{\ddagger}$	&2.84E-01$^{\ddagger}$	&\textbf{2.20E-01}$^{\ddagger}$\\
$F_{wht}$	&3.73E+02	&2.00E+02$^{\ddagger}$	&2.49E+02$^{\ddagger}$	&2.27E+02$^{\ddagger}$	&\textbf{1.93E+02}$^{\ddagger}$\\
$F_{pn1}$	&2.07E-03	&\textbf{1.57E-32}$^{\ddagger}$	&\textbf{1.57E-32}$^{\ddagger}$	&2.07E-03$^{\approx}$	&\textbf{1.57E-32}$^{\ddagger}$\\
$F_{pn2}$	&4.82E-28	&\textbf{1.35E-32}$^{\ddagger}$	&\textbf{1.35E-32}$^{\ddagger}$	&\textbf{1.35E-32}$^{\ddagger}$	&\textbf{1.35E-32}$^{\ddagger}$\\
$F_{1}$	&	6.48E-30	&\textbf{0.00E+00}$^{\ddagger}$	&\textbf{0.00E+00}$^{\ddagger}$	&\textbf{0.00E+00}$^{\ddagger}$	&\textbf{0.00E+00}$^{\ddagger}$\\
$F_{2}$	&	1.05E-02	&1.10E-03$^{\ddagger}$	&\textbf{2.77E-04}$^{\ddagger}$	&1.40E-03$^{\ddagger}$	&1.23E-02$^{\dagger}$\\
$F_{3}$	&	\textbf{2.17E+06}	&3.04E+06$^{\dagger}$	&2.99E+06$^{\dagger}$	&3.25E+06$^{\dagger}$	&3.61E+06$^{\dagger}$\\
$F_{4}$	&	1.79E+03	&4.98E+02$^{\ddagger}$	&\textbf{3.58E+02}$^{\ddagger}$	&4.02E+02$^{\ddagger}$	&5.51E+02$^{\ddagger}$\\
$F_{5}$	&	3.17E+03	&2.44E+03$^{\ddagger}$	&1.98E+03$^{\ddagger}$	&2.11E+03$^{\ddagger}$	&\textbf{1.94E+03}$^{\ddagger}$\\
$F_{6}$	&	2.78E+01	&1.57E+01$^{\ddagger}$	&\textbf{9.91E+00}$^{\ddagger}$	&1.23E+01$^{\ddagger}$	&1.59E+01$^{\ddagger}$\\
$F_{7}$	&	\textbf{1.34E-03}	&2.22E-03$^{\dagger}$	&5.79E-03$^{\dagger}$	&3.73E-03$^{\dagger}$	&2.49E-03$^{\dagger}$\\
$F_{8}$	&	\textbf{2.09E+01}	&\textbf{2.09E+01}$^{\approx}$	&\textbf{2.09E+01}$^{\approx}$	&\textbf{2.09E+01}$^{\approx}$	&\textbf{2.09E+01}$^{\approx}$\\
$F_{9}$	&	2.84E+01	&1.30E+01$^{\ddagger}$	&2.74E+01$^{\ddagger}$	&1.28E+01$^{\ddagger}$	&\textbf{6.81E+00}$^{\ddagger}$\\
$F_{10}$	&	1.69E+02	&	1.21E+02$^{\ddagger}$	&	1.31E+02$^{\ddagger}$	&	1.18E+02$^{\ddagger}$	&\textbf{1.03E+02}$^{\ddagger}$\\
$\ddagger/\approx/\dagger$ &-  &17/1/2  &17/1/2  &16/2/2  &16/1/3 \\
p-value &-   &8.97E-03  &1.00E-02   &1.00E-02   &1.87E-02 \\
Avg. rank   &4.35  &2.78  &2.88   &2.58  &\textbf{2.43}    \\
\bottomrule
\end{tabular*}%}
\end{threeparttable}
\end{center}
\end{table}

To further discuss the convergence speed of the four randomness-enhanced CS algorithms, several test problems (namely $F_{sph}$, $F_{grw}$, $F_{1}$ and $F_{10}$) at $D=30$ are selected to plot the convergence curves of the averages of the function error values within Max$\_$FEs over 50 independent runs, which are presented in Figure~\ref{figcompare}. From Figure~\ref{figcompare}, it can be observed that CSML, CSP, CSC, and CSW converge outstandingly faster than CS according to the convergence curves. In summary, it can be concluded that the standard CS algorithm can be improved by integrating different heavy-tailed probability distributions rather than L\'{e}vy distribution into it.

\begin{figure}[htbp]
\centering
\subfigure[]{
%\subfigure[pic1.]{
\begin{minipage}[t]{0.3\linewidth}
\centering
\includegraphics[width=2.in]{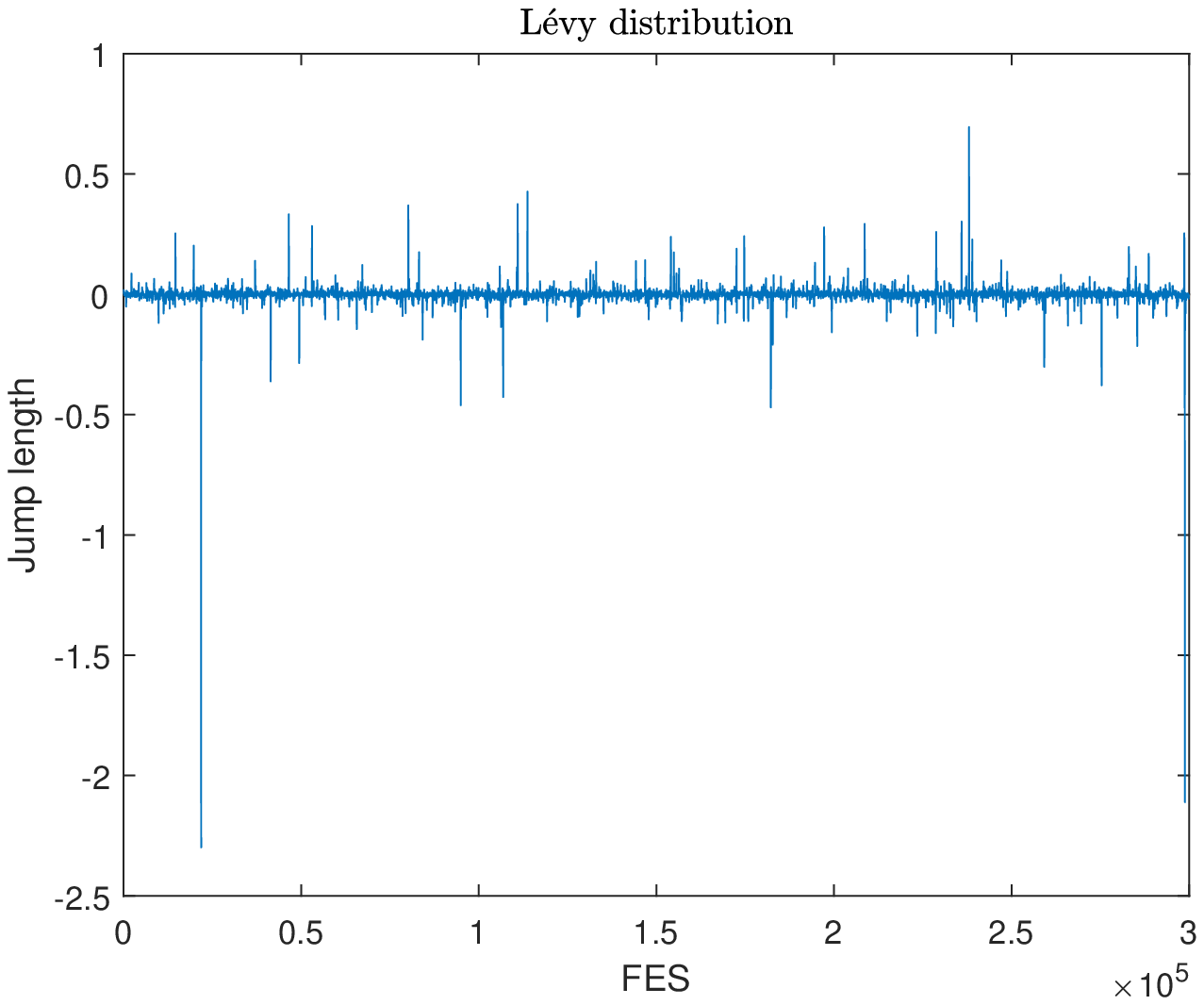}
\end{minipage}\label{figproL}
}%
\subfigure[]{
%\subfigure[pic2.]{
\begin{minipage}[t]{0.3\linewidth}
\centering
\includegraphics[width=2.in]{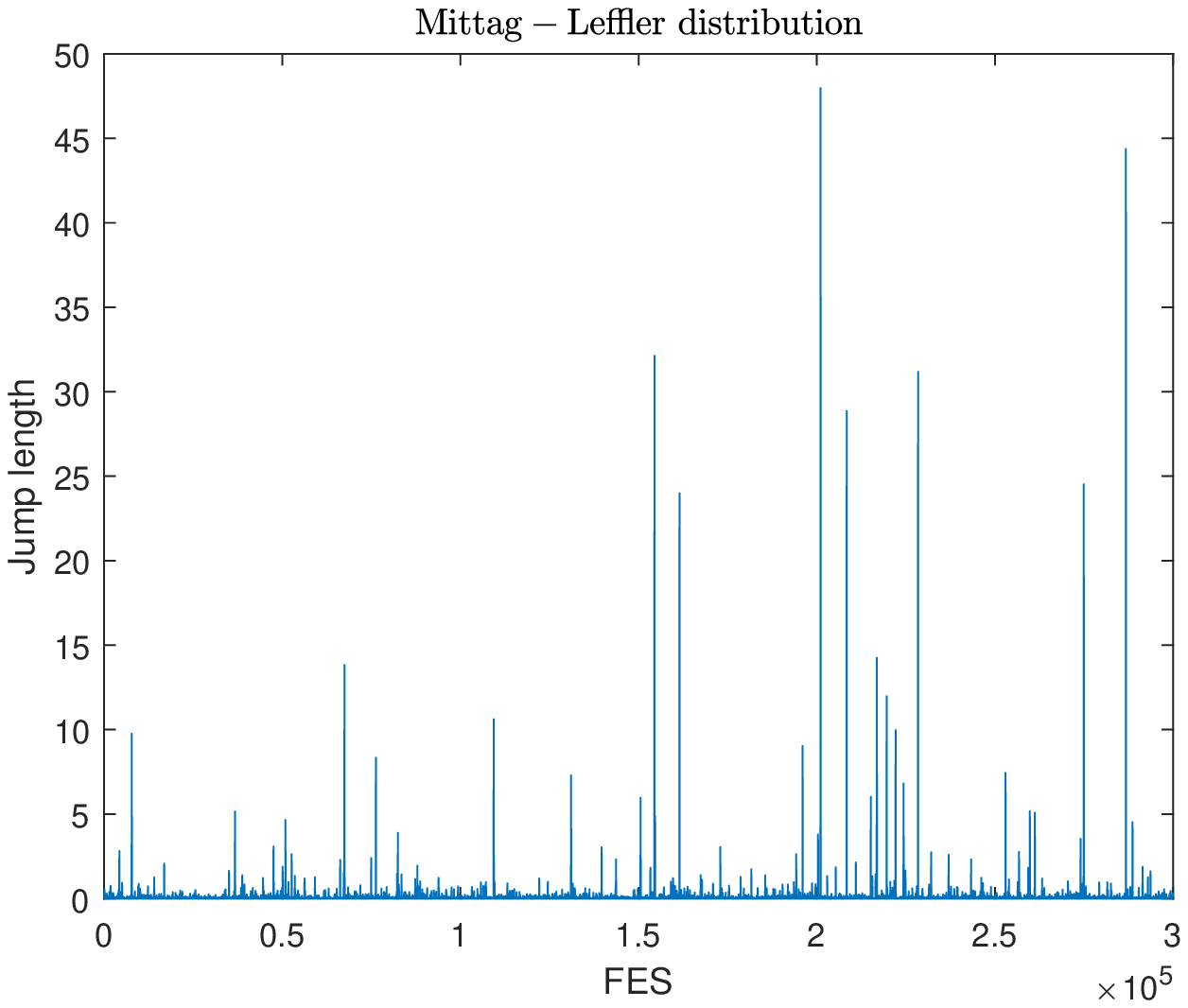}
%\caption{fig2}
\end{minipage}%
}%
\\
\subfigure[]{
\begin{minipage}[t]{0.3\linewidth}
\centering
\includegraphics[width=2.in]{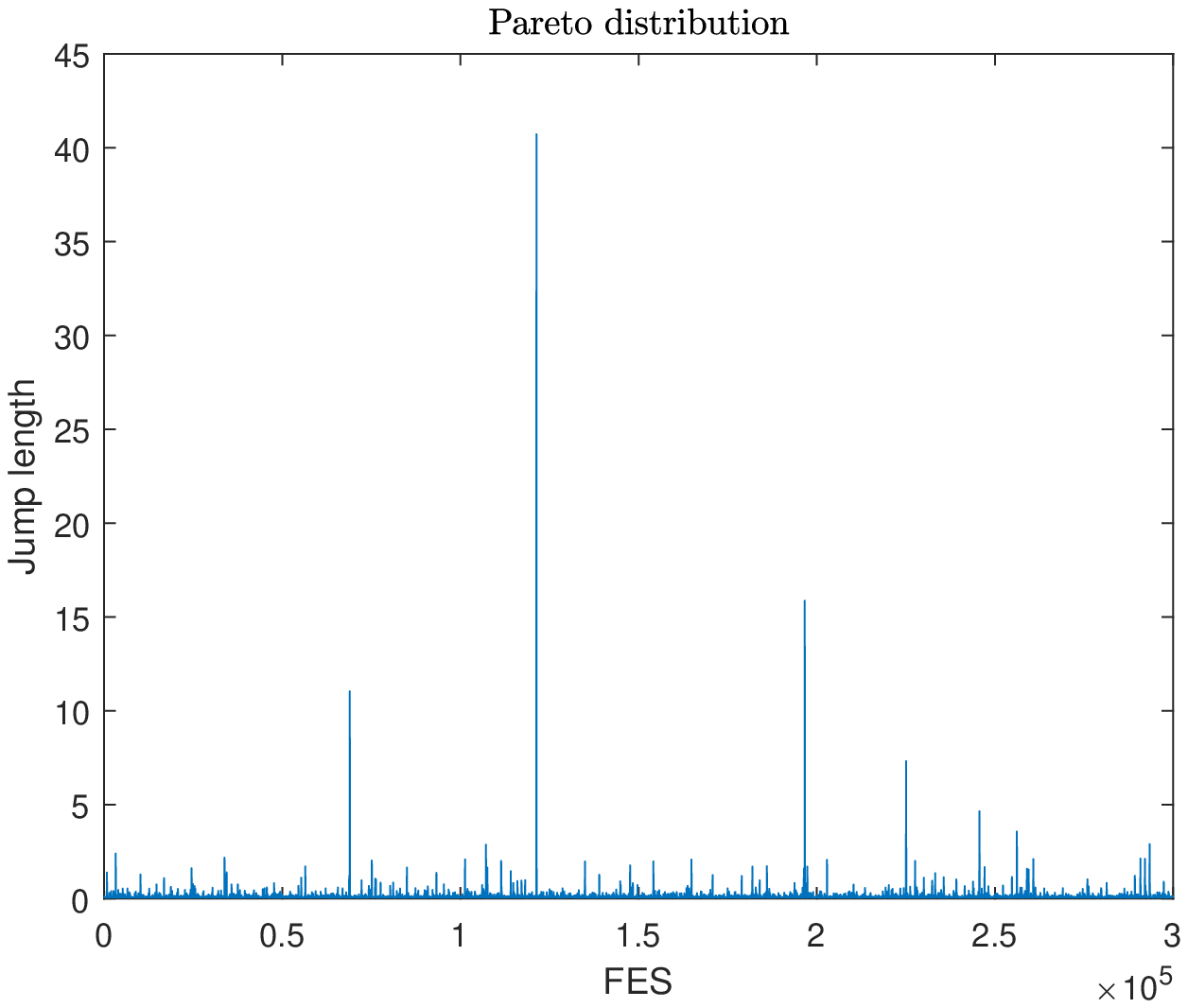}
%\caption{fig1}
\end{minipage}%
}%
\subfigure[]{
%\subfigure[pic2.]{
\begin{minipage}[t]{0.3\linewidth}
\centering
\includegraphics[width=2.in]{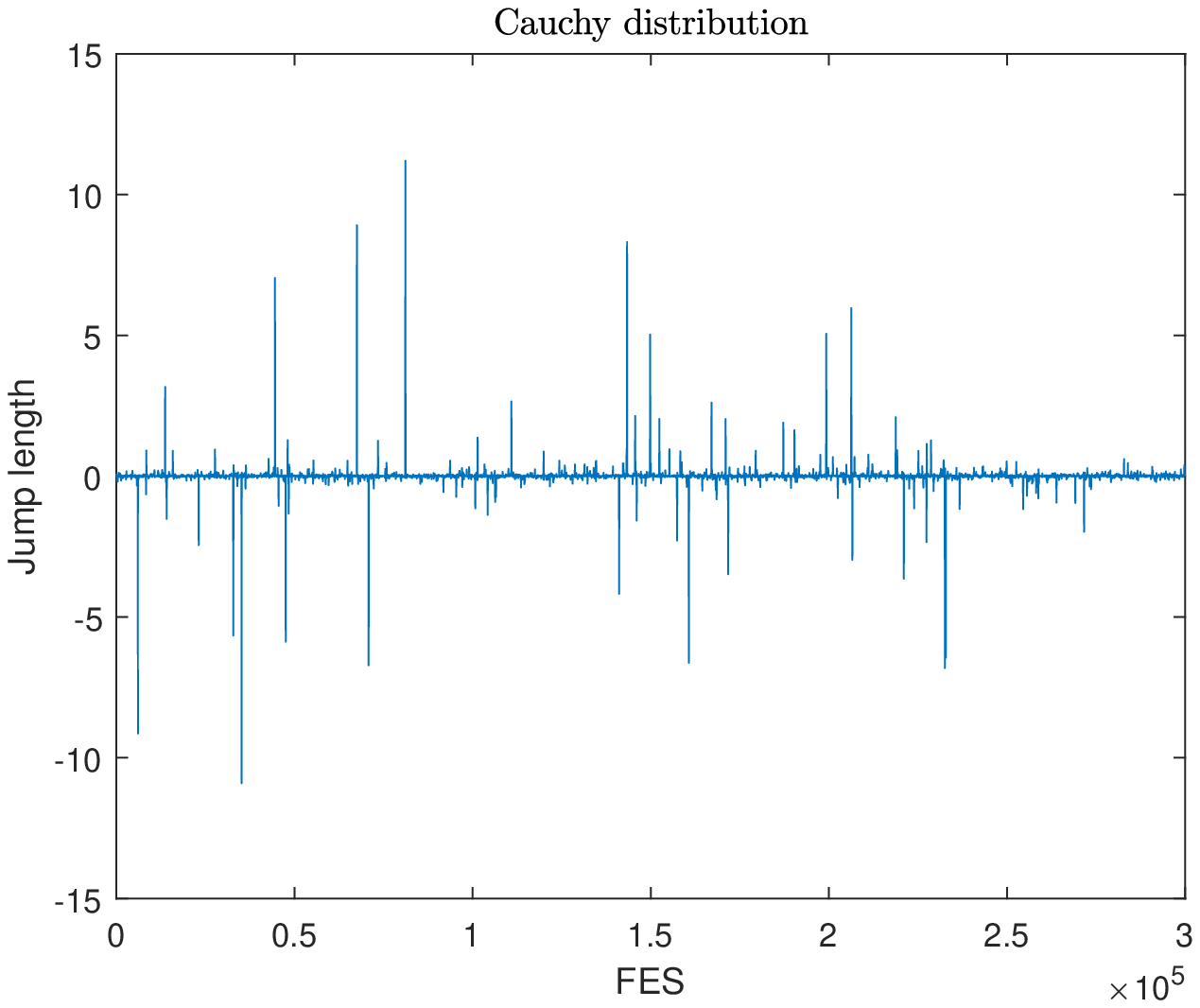}
%\caption{fig2
\end{minipage}%
}%
\subfigure[]{
%\subfigure[pic2.]{
\begin{minipage}[t]{0.3\linewidth}
\centering
\includegraphics[width=2.in]{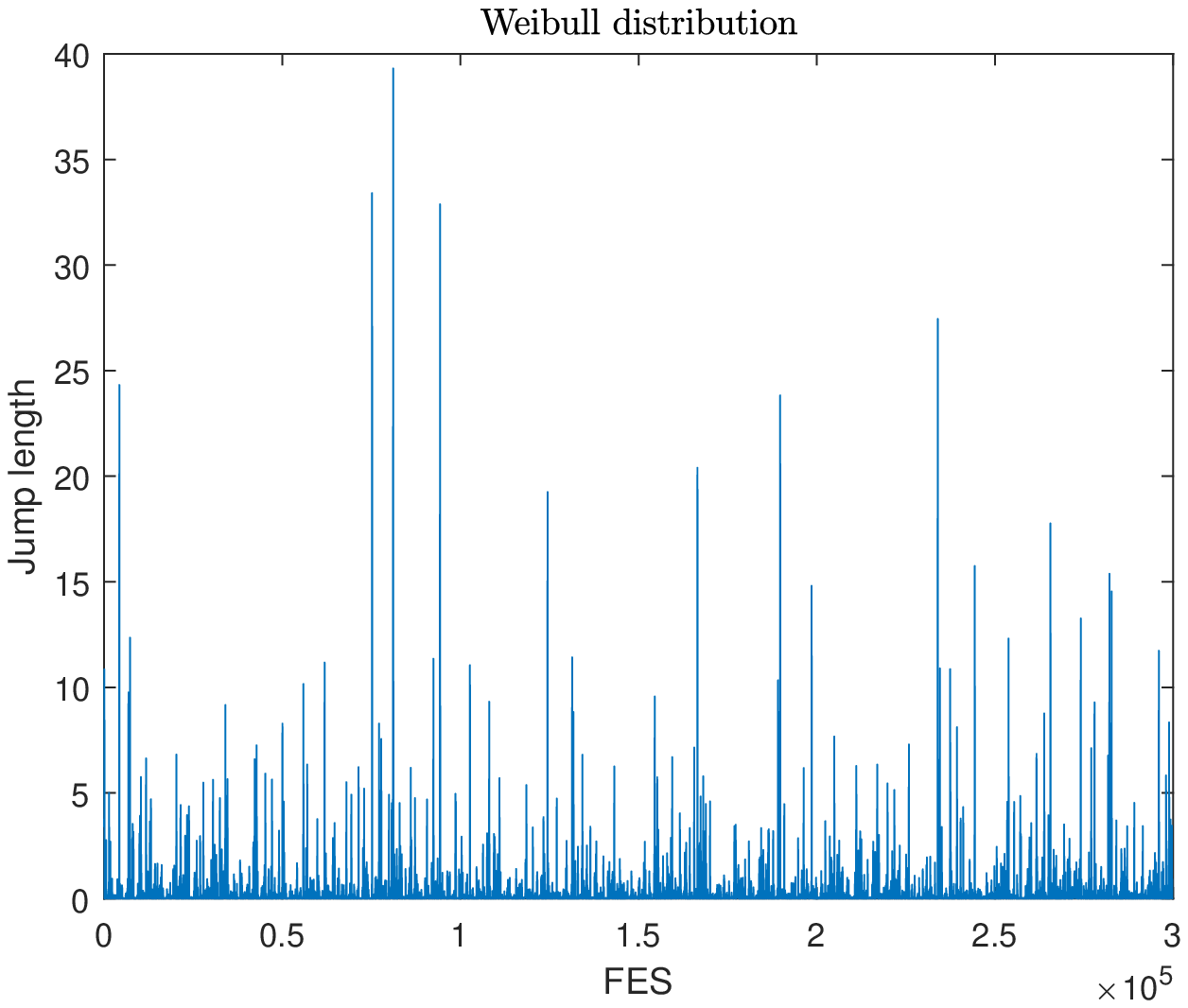}
%\caption{fig2
\end{minipage}%
}%
\caption{Jump lengths of CS, CSML, CSP, CSC and CSW.}
\label{figsteplength}
\end{figure}

Besides, to analyze the reasons for different performances among the four proposed randomness-enhanced CS algorithms, the jump lengths of CS, CSML, CSP, CSC, and CSW (namely, $\alpha\otimes {\rm L\acute{e}vy}(\lambda)$, $\alpha\otimes {\rm MittagLeffler}(\beta,\gamma)$, $\alpha\otimes {\rm Pareto}(b,a)$, $\alpha\otimes {\rm Cauchy}(\mu,\sigma)$, and $\alpha\otimes {\rm Weibull}(\xi,\kappa)$) are depicted in Figure~\ref{figsteplength}, where the parameters are given in Table~\ref{Paraforalgs} and the scaling factor is set to $0.01$. From Figure~\ref{figsteplength}, it can be observed that (1) L{\'e}vy distribution and Cauchy distribution are one-sided distribution where all the random numbers are positive, and the other three distributions are two-sided; (2) large steps frequently take place for all distributions; (3) since the tail of Weibull distribution is the lightest, the extreme large steps (compared with its mean) are less likely to happen.

\subsection{Scalability Study}
In this section, a scalability study comparing with the standard CS algorithm is conducted in order to study the effect of problem size on the performance of the four proposed randomness-enhanced CS algorithms. We carry out experiments on the 20 benchmark functions with dimension $D$ set to 10 and 50. When $D=10$, the population size is chosen as $NP=30$; meanwhile, when $D=50$, the population size is selected as $NP=D$. All the other control parameters are kept unchanged. The experimental results achieved by CS and four proposed randomness-enhanced CS algorithms at $D=10$ and $D=50$ are listed in Tables~\ref{Compare10D} and \ref{Compare50D}, respectively, and the results of the Wilcoxon signed-rank test are also given in the tables.

\begin{table}[htbp]
\renewcommand\arraystretch{0.8}
\caption{Comparisons between CS and four randomness-enhanced CS algorithms at $D=10$.}\label{Compare10D}
\begin{center}
%\resizebox{\textwidth}{!}{
\begin{threeparttable}
\begin{tabular*}{\hsize}{@{}@{\extracolsep{\fill}}lllllllll@{}} \toprule
%\begin{tabular}{lllllllll} \toprule
Fun &CS  &CSML &CSP &CSC &CSW  \cr \midrule			
$F_{sph}$	&	4.87E-26	&	3.39E-31$^{\ddagger}$	&	\textbf{4.21E-48}$^{\ddagger}$	&	2.04E-46$^{\ddagger}$	&	2.48E-46$^{\ddagger}$	\\
$F_{ros}$	&	9.63E-01	&	3.02E+01$^{\dagger}$	&	\textbf{1.12E-01}$^{\ddagger}$	&	1.75E-01$^{\ddagger}$	&	2.97E-01$^{\ddagger}$	\\
$F_{ack}$	&	4.16E-11	&	2.50E-14$^{\ddagger}$	&	\textbf{4.37E-15}$^{\ddagger}$	&	4.44E-15$^{\ddagger}$	&	4.44E-15$^{\ddagger}$	\\
$F_{grw}$	&	3.44E-02	&	\textbf{0.00E+00}$^{\ddagger}$	&	2.22E-02$^{\ddagger}$	&	2.07E-02$^{\ddagger}$	&	1.44E-02$^{\ddagger}$	\\
$F_{ras}$	&	3.00E+00	&	6.93E+01$^{\dagger}$	&	2.25E+00$^{\ddagger}$	&	2.95E-01$^{\ddagger}$	&	\textbf{2.26E-09}$^{\ddagger}$	\\
$F_{sch}$	&	6.72E+01	&	3.80E+03$^{\dagger}$	&	1.38E+01$^{\ddagger}$	&	6.91E-03$^{\ddagger}$	&	\textbf{1.27E-04}$^{\ddagger}$	\\
$F_{sal}$	&	1.04E-01	&	4.78E-01$^{\dagger}$	&	\textbf{9.99E-02}$^{\ddagger}$	&	\textbf{9.99E-02}$^{\ddagger}$	&	\textbf{9.99E-02}$^{\ddagger}$	\\
$F_{wht}$	&	2.40E+01	&	9.89E+02$^{\dagger}$	&	1.54E+01$^{\ddagger}$	&	1.01E+01$^{\ddagger}$	&	\textbf{5.72E+00}$^{\ddagger}$	\\
$F_{pn1}$	&	1.96E-16	&	2.01E-28$^{\ddagger}$	&	\textbf{4.71E-32}$^{\ddagger}$	&	\textbf{4.71E-32}$^{\ddagger}$	&	\textbf{4.71E-32}$^{\ddagger}$	\\
$F_{pn2}$	&	4.86E-23	&	9.17E-30$^{\ddagger}$	&	\textbf{1.35E-32}$^{\ddagger}$	&	\textbf{1.35E-32}$^{\ddagger}$	&	\textbf{1.35E-32}$^{\ddagger}$	\\
$F_{1}$	&	4.13E-26	&	\textbf{0.00E+00}$^{\ddagger}$	&	\textbf{0.00E+00}$^{\ddagger}$	&	\textbf{0.00E+00}$^{\ddagger}$	&	\textbf{0.00E+00}$^{\ddagger}$	\\
$F_{2}$	&	8.16E-14	&	3.73E+02$^{\dagger}$	&	\textbf{1.33E-21}$^{\ddagger}$	&	4.54E-19$^{\ddagger}$	&	1.51E-16$^{\ddagger}$	\\
$F_{3}$	&	\textbf{2.08E+02}	&	1.70E+07$^{\dagger}$	&	7.20E+02$^{\dagger}$	&	6.78E+02$^{\dagger}$	&	8.31E+02$^{\dagger}$	\\
$F_{4}$	&	1.01E-05	&	1.96E+04$^{\dagger}$	&	\textbf{1.46E-09}$^{\ddagger}$	&	1.20E-08$^{\ddagger}$	&	4.82E-08$^{\ddagger}$	\\
$F_{5}$	&	9.30E-05	&	6.82E+03$^{\dagger}$	&	\textbf{6.13E-10}$^{\ddagger}$	&	5.11E-09$^{\ddagger}$	&	9.27E-09$^{\ddagger}$	\\
$F_{6}$	&	9.78E-01	&	4.11E+01$^{\dagger}$	&	6.38E-01$^{\ddagger}$	&	3.48E-01$^{\ddagger}$	&	\textbf{2.69E-01}$^{\ddagger}$	\\
$F_{7}$	&	5.33E-02	&	\textbf{1.07E-03}$^{\ddagger}$	&	5.91E-02$^{\dagger}$	&	4.72E-02$^{\ddagger}$	&	4.39E-02$^{\ddagger}$	\\
$F_{8}$	&	2.04E+01	&	2.11E+01$^{\dagger}$	&	2.04E+01$^{\approx}$	&	2.04E+01$^{\approx}$	&	\textbf{2.03E+01}$^{\ddagger}$	\\
$F_{9}$	&	2.75E+00	&	7.37E+01$^{\dagger}$	&	1.80E+00$^{\ddagger}$	&	1.79E-01$^{\ddagger}$	&	\textbf{2.35E-10}$^{\ddagger}$	\\
$F_{10}$	&	1.99E+01	&	2.89E+02$^{\dagger}$	&	1.63E+01$^{\ddagger}$	&	1.59E+01$^{\ddagger}$	&	\textbf{1.43E+01}$^{\ddagger}$	\\
$\ddagger/\approx/\dagger$ &-  &7/0/13  &17/1/2  &18/1/1  &19/1/0 \\
Avg. rank   &4.10  &4.28  &2.28   &2.25  &\textbf{2.10}    \\
\bottomrule
\end{tabular*}%}
\end{threeparttable}
\end{center}
\end{table}

According to Table~\ref{Compare10D}, CSML, CSP, CSC, and CSW are significantly better than CS on 7, 17, 18 and 19 test functions, similar to CS on 0, 1, 1 and 1 test functions, and worse than CS on 13, 2, 1 and 0 test functions, respectively. The comprehensive ranking orders in the case of $D=10$ are CSW, CSC, CSP, CS, and CSML in descending manner. The results show that the performance improvement of using different heavy-tailed probability distributions persists expect CSML when the problem dimension reduces to 10. In the case of $D=50$, it can be observed from Table~\ref{Compare50D} that CSML, CSP, CSC and CSW perform better than CS on 16, 14, 16 and 16 test functions, to CS on 1, 1, 1 and 1 test functions, and worse than CS on 3, 5, 3 and 3 test functions, respectively. Meanwhile, the corresponding comprehensive ranking orders when $D=50$ are CSC, CSML, CSP, CSW and CS. In general, we can draw conclusions that the advantages of four randomness-enhanced CS algorithms over the standard CS are overall stable when the problem dimension increases, except CSML which deteriorates to a certain extent when $D$ set to 10. Furthermore, regarding to the different comprehensive ranking orders obtained at every dimension, it is pointed out that CS with L\'{e}vy flights seems not the optimal randomness when compared with those using different heavy-tailed probability distributions in CS.

%% table 4
\begin{table}[htbp]
\renewcommand\arraystretch{0.8}
\caption{Comparisons between CS and four randomness-enhanced CS algorithms at $D=50$.}\label{Compare50D}
\begin{center}
%\resizebox{\textwidth}{!}{
\begin{threeparttable}
\begin{tabular*}{\hsize}{@{}@{\extracolsep{\fill}}lllllllll@{}} \toprule
%\begin{tabular}{lllllllll} \toprule
Fun &CS  &CSML &CSP &CSC &CSW  \cr \midrule			
$F_{sph}$	&	3.79E-17	&	3.47E-31$^{\ddagger}$	&	\textbf{7.41E-36}$^{\ddagger}$	&	5.75E-32$^{\ddagger}$	&	2.73E-24$^{\ddagger}$	\\
$F_{ros}$	&	4.22E+01	&	3.07E+01$^{\ddagger}$	&	\textbf{2.82E+01}$^{\ddagger}$	&	2.99E+01$^{\ddagger}$	&	3.41E+01$^{\ddagger}$	\\
$F_{ack}$	&	2.85E-02	&	2.43E-14$^{\ddagger}$	&	\textbf{2.05E-14}$^{\ddagger}$	&	\textbf{2.05E-14}$^{\ddagger}$	&	7.40E-13$^{\ddagger}$	\\
$F_{grw}$	&	1.93E-10	&	\textbf{0.00E+00}$^{\ddagger}$	&	\textbf{0.00E+00}$^{\ddagger}$	&	\textbf{0.00E+00}$^{\ddagger}$	&	\textbf{0.00E+00}$^{\ddagger}$	\\
$F_{ras}$	&	8.44E+01	&	\textbf{6.80E+01}$^{\ddagger}$	&	8.69E+01$^{\dagger}$	&	7.54E+01$^{\ddagger}$	&	7.37E+01$^{\ddagger}$	\\
$F_{sch}$	&	4.87E+03	&	4.14E+03$^{\ddagger}$	&	6.05E+03$^{\dagger}$	&	4.38E+03$^{\ddagger}$	&	\textbf{2.38E+03}$^{\ddagger}$	\\
$F_{sal}$	&	6.69E-01	&	4.68E-01$^{\ddagger}$	&	4.87E-01$^{\ddagger}$	&	4.22E-01$^{\ddagger}$	&	\textbf{3.68E-01}$^{\ddagger}$	\\
$F_{wht}$	&	1.36E+03	&	\textbf{9.58E+02}$^{\ddagger}$	&	1.21E+03$^{\ddagger}$	&	1.09E+03$^{\ddagger}$	&	1.13E+03$^{\ddagger}$	\\
$F_{pn1}$	&	8.13E-03	&	6.74E-28$^{\ddagger}$	&	1.04E-27$^{\ddagger}$	&	\textbf{7.21E-30}$^{\ddagger}$	&	1.47E-23$^{\ddagger}$	\\
$F_{pn2}$	&	3.25E-14	&	1.02E-29$^{\ddagger}$	&	\textbf{1.57E-32}$^{\ddagger}$	&	1.44E-30$^{\ddagger}$	&	2.01E-23$^{\ddagger}$	\\
$F_{1}$	&	1.40E-16	&	\textbf{0.00E+00}$^{\ddagger}$	&	\textbf{0.00E+00}$^{\ddagger}$	&	\textbf{0.00E+00}$^{\ddagger}$	&	3.57E-24$^{\ddagger}$	\\
$F_{2}$	&	2.34E+02	&	3.57E+02$^{\dagger}$	&	\textbf{1.86E+02}$^{\ddagger}$	&	4.49E+02$^{\dagger}$	&	8.59E+02$^{\dagger}$	\\
$F_{3}$	&	\textbf{8.53E+06}	&	1.66E+07$^{\dagger}$	&	1.47E+07$^{\dagger}$	&	1.83E+07$^{\dagger}$	&	1.85E+07$^{\dagger}$	\\
$F_{4}$	&	2.72E+04	&	1.99E+04$^{\ddagger}$	&	\textbf{1.72E+04}$^{\ddagger}$	&	1.91E+04$^{\ddagger}$	&	1.88E+04$^{\ddagger}$	\\
$F_{5}$	&	1.06E+04	&	6.95E+03$^{\ddagger}$	&	6.49E+03$^{\ddagger}$	&	6.65E+03$^{\ddagger}$	&	\textbf{6.30E+03}$^{\ddagger}$	\\
$F_{6}$	&	6.38E+01	&	3.90E+01$^{\ddagger}$	&	4.15E+01$^{\ddagger}$	&	\textbf{3.63E+01}$^{\ddagger}$	&	4.43E+01$^{\ddagger}$	\\
$F_{7}$	&	\textbf{1.30E-03}	&	1.81E-03$^{\dagger}$	&	3.56E-03$^{\dagger}$	&	2.43E-03$^{\dagger}$	&	3.64E-03$^{\dagger}$	\\
$F_{8}$	&	\textbf{2.11E+01}	&	\textbf{2.11E+01}$^{\approx}$	&	\textbf{2.11E+01}$^{\approx}$	&	\textbf{2.11E+01}$^{\approx}$	&	\textbf{2.11E+01}$^{\approx}$	\\
$F_{9}$	&	1.24E+02	&	7.04E+01$^{\ddagger}$	&	1.27E+02$^{\dagger}$	&	7.47E+01$^{\ddagger}$	&	\textbf{6.50E+01}$^{\ddagger}$	\\
$F_{10}$	&	3.87E+02	&	2.87E+02$^{\ddagger}$	&	3.13E+02$^{\ddagger}$	&	2.85E+02$^{\ddagger}$	&	\textbf{2.69E+02}$^{\ddagger}$	\\
$\ddagger/\approx/\dagger$ &-  &16/1/3  &14/1/5  &16/1/3  &16/1/3 \\
Avg. rank   &4.10  &2.58  &2.70  &\textbf{2.55}   &3.08   \\
\bottomrule
\end{tabular*}%}
\end{threeparttable}
\end{center}
\end{table}

\subsection{Comparison with Other Optimization Algorithms}
In order to demonstrate the superiority of the four proposed randomness-enhanced CS algorithms, we compare the performance of CSML, CSP, CSC and CSW with several classical state-of-the-art optimization algorithms, namely ABC\cite{karaboga2007powerful}, DE\cite{storn1997differential}, FA\cite{yang2009firefly}, FPA\cite{yang2012flower} and PSO\cite{kennedy1995particle}, by conducting numerical experiments on the 20 benchmark functions at dimension $D=30$. In our experimental study, Max$\_$FEs set to $10,000\times D$ is taken as the termination criterion, and population size is set to 30. For ABC, the number of food sources $SN=30$, maximum number of trial for abandoning a source $limit=100$; for PSO, inertia weight defined as $\omega^{t+1}=\omega^{t+1}*0.99$ and $t$ denotes iteration number, acceleration constants $c1=1.5$, $c2=2.0$. The MATLAB source codes of FA and FPA are obtained from \cite{yang2014nature}. And the results of DE are taken from the literature \cite{noman2008accelerating} which has the same termination criterion. The comparative simulation results of all the optimization algorithms are listed in Table~\ref{Comparewithotheralgs}. Additionally, Table~\ref{Comparewithotheralgs} records the statistical results obtained by both of the Wilcoxon signed-rank test and the Friedman test for 20 benchmark functions at dimension $D=30$.

The experimental results in Table~\ref{Comparewithotheralgs} clearly demonstrate that all of the four proposed randomness-enhanced CS algorithms perform better on the majority of benchmark functions. More specifically, CSW is overall the best, CSC is the second best, CSML is the third best, and CSP is the fourth best followed by DE, PSO, ABC, FPA, and FA. This suggests that the proposed randomness-enhanced CS algorithms are also highly competitive when compared with other optimization algorithms.

\begin{table}[htbp]
\caption{Comparisons of four randomness-enhanced CS algorithms with other optimization algorithms at $D=30$.}\label{Comparewithotheralgs}
\renewcommand\arraystretch{0.8}
\begin{center}
\resizebox{\textwidth}{!}{
\begin{threeparttable}
%\begin{tabular*}{\hsize}{@{}@{\extracolsep{\fill}}llllllllllll@{}} \toprule
\begin{tabular}{llllllllllll} \toprule
Fun &CSML &CSP &CSC &CSW &ABC &DE &FA &FPA &PSO  \cr \midrule			
$F_{sph}$	&	4.90E-54	&	4.74E-59	&	1.17E-57	&	4.40E-51	&	\textbf{6.28E-157}	&	5.73E-17	&	2.90E+03	&	8.65E-04	&	2.05E-06	\\
$F_{ros}$	&	5.22E+00	&	3.10E+00	&	\textbf{2.74E+00}	&	8.62E+00	&	3.00E+02	&	5.20E+01	&	7.91E+07	&	1.16E+02	&	3.99E+01	\\
$F_{ack}$	&	1.06E-14	&	1.07E-14	&	9.56E-15	&	\textbf{8.28E-15}	&	2.51E-01	&	1.37E-09	&	1.07E+01	&	1.57E+00	&	1.90E+01	\\
$F_{grw}$	&	\textbf{0.00E+00}	&	\textbf{0.00E+00}	&	\textbf{0.00E+00}	&	\textbf{0.00E+00}	&	5.33E+00	&	2.66E-03	&	2.71E+01	&	2.39E-02	&	9.76E-03	\\
$F_{ras}$	&	1.38E+01	&	1.88E+01	&	1.49E+01	&	8.34E+00	&	\textbf{1.99E-01}	&	2.55E+01	&	2.20E+02	&	4.05E+01	&	1.21E+02	\\
$F_{sch}$	&	5.37E+02	&	1.32E+03	&	4.80E+02	&	\textbf{3.56E+01}	&	1.65E+02	&	4.90E+02	&	1.00E+04	&	3.49E+03	&	1.13E+04	\\
$F_{sal}$	&	2.96E-01	&	3.00E-01	&	2.84E-01	&	\textbf{2.20E-01}	&	2.98E+00	&	2.52E-01	&	5.60E+00	&	1.74E+00	&	8.94E-01	\\
$F_{wht}$	&	2.00E+02	&	2.49E+02	&	2.27E+02	&	\textbf{1.93E+02}	&	6.79E+03	&	3.10E+02	&	1.30E+13	&	5.14E+04	&	7.04E+02	\\
$F_{pn1}$	&	\textbf{1.57E-32}	&	\textbf{1.57E-32}	&	2.07E-03	&	\textbf{1.57E-32}	&	8.18E+06	&	4.56E-02	&	4.58E+02	&	4.41E-01	&	4.09E-01	\\
$F_{pn2}$	&	\textbf{1.35E-32}	&	\textbf{1.35E-32}	&	\textbf{1.35E-32}	&	\textbf{1.35E-32}	&	1.38E-15	&	1.44E-01	&	3.94E+05	&	2.47E+00	&	7.69E-01	\\
$F_{1}$	&	\textbf{0.00E+00}	&	\textbf{0.00E+00}	&	\textbf{0.00E+00}	&	\textbf{0.00E+00}	&	3.81E+03	&	3.87E-14	&	6.39E+04	&	9.32E-03	&	1.08E-05	\\
$F_{2}$	&	1.10E-03	&	2.77E-04	&	1.40E-03	&	1.23E-02	&	4.19E+03	&	8.50E-02	&	9.45E+04	&	8.61E+00	&	\textbf{3.79E-14}	\\
$F_{3}$	&	3.04E+06	&	2.99E+06	&	3.25E+06	&	3.61E+06	&	2.90E+07	&	3.63E+06	&	1.71E+09	&	\textbf{1.00E+05}	&	1.53E+06	\\
$F_{4}$	&	4.98E+02	&	3.58E+02	&	4.02E+02	&	5.51E+02	&	1.26E+05	&	\textbf{5.54E+01}	&	1.23E+05	&	5.16E+03	&	1.11E+03	\\
$F_{5}$	&	2.44E+03	&	1.98E+03	&	2.11E+03	&	1.94E+03	&	1.45E+04	&	\textbf{1.08E+03}	&	4.28E+04	&	1.74E+03	&	5.66E+03	\\
$F_{6}$	&	1.57E+01	&	\textbf{9.91E+00}	&	1.23E+01	&	1.59E+01	&	1.62E+02	&	6.67E+01	&	2.85E+10	&	2.69E+03	&	4.17E+01	\\
$F_{7}$	&	\textbf{2.22E-03}	&	5.79E-03	&	3.73E-03	&	2.49E-03	&	4.70E+03	&	7.59E-03	&	1.22E+04	&	4.70E+03	&	2.71E+03	\\
$F_{8}$	&	\textbf{2.09E+01}	&	\textbf{2.09E+01}	&	\textbf{2.09E+01}	&	\textbf{2.09E+01}	&	2.12E+01	&	2.09E+01	&	2.12E+01	&	2.10E+01	&	\textbf{2.09E+01}	\\
$F_{9}$	&	1.30E+01	&	2.74E+01	&	1.28E+01	&	6.81E+00	&	\textbf{1.65E-01}	&	2.43E+01	&	3.89E+02	&	9.23E+01	&	1.21E+02	\\
$F_{10}$	&	1.21E+02	&	1.31E+02	&	1.18E+02	&	1.03E+02	&	4.58E+02	&	\textbf{7.33E+01}	&	6.42E+02	&	1.91E+02	&	2.23E+02	\\
Avg. rank   &3.85  &3.95  &3.60  &\textbf{3.35}   &7.20 &5.50 &9.78 &7.58 &6.95   \\
\bottomrule
\end{tabular}
\end{threeparttable}}
\end{center}
\end{table}

\subsection{Application to Parameter Identification of Fractional-Order Chaotic Systems}
In this section, the four proposed randomness-enhanced CS algorithms (namely, CSML, CSP, CSC, and CSW) are applied to identify unknown parameters of fractional-order chaotic systems, which is a critical issue in chaos control and synchronization. Our main task of this section is to further demonstrate that improving CS with different heavy-tailed probability distributions can also effectively tackle the real-world complex optimization problems besides the benchmark problems. In fact, by using a non-Lyapunov way according to problem formulation suggested in \cite{gao2014inversion}, the nonlinear function optimization can be converted to from parameter identification of uncertain fractional-order chaotic systems.

\begin{figure}[htbp]
\begin{center}
\includegraphics[width=3.2in]{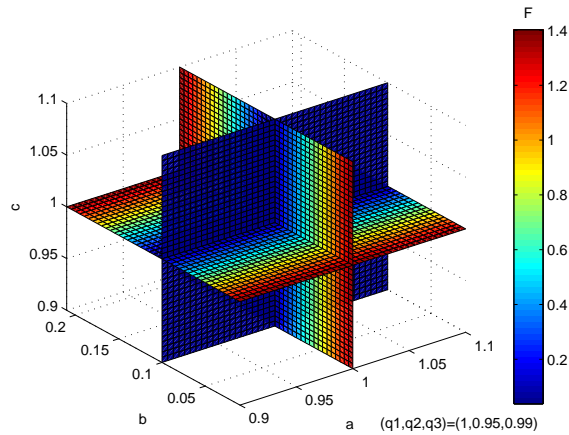}
\end{center}
\caption{Distribution of the objective function values for system (\ref{FOFS}).}
\label{figDIS}
\end{figure}

In the numerical simulation, the fractional-order financial system \cite{Chen2008Nonlinear} under the Caputo definition is taken for example, which can be described as
\begin{eqnarray} \label{FOFS}
\left\{\begin{array}{l}
_{0}D^{q_{1}}_{t}x(t)=z(t)+x(t)(y(t)-a),\\
_{0}D^{q_{2}}_{t}y(t)=1-by(t)-x^{2}(t),\\
_{0}D^{q_{3}}_{t}z(t)=-x(t)-cz(t),\\
\end{array}
\right.
\end{eqnarray}
where $q_{1},q_{2},q_{3}$ and $a,b,c$ are fractional orders and systematic parameters. When $(q_{1},q_{2},q_{3})=(1,0.95,0.99)$, $(a,b,c)=(1,0.1,1)$, and initial point $(x_{0},y_{0},z_{0})=(2,-1,1)$, the system above is chaotic. Figure~\ref{figDIS} depicts the distribution figure of system (\ref{FOFS}) for the objective function values.

\begin{table}[htbp]
\caption{Statistical results of different methods for system (\ref{FOFS}).}
\begin{center}
\label{tabAPP}
\begin{scriptsize}
\begin{tabular*}{\hsize}{@{}@{\extracolsep{\fill}}lllllllll@{}} \toprule
Method   &CS   &CSML  &CSP  &CSC  & CSW    \\ \midrule
% Mean     &     &         \\
$a$     &0.999999825481796   &0.999999979386471    &\textbf{1.000000001165006}    &0.999999930875086       &0.999999994619958    \\
$\frac{|a-1.00|}{1.00}$     &1.75E-07    &2.28E-08       &\textbf{1.17E-09}       &6.91E-08       &5.38E-09     \\
$b$    &0.100000078306700    &0.100000006492360  &\textbf{0.099999999732393}       &0.100000038684769       &0.100000001325757      \\
$\frac{|b-0.10|}{0.10}$     &7.83E-07                 &1.12E-07              &\textbf{2.68E-09}       &3.87E-07       &2.06E-08      \\
$c$    &1.000000126069434  &0.999999979588057  &\textbf{0.999999995606294}       &0.999999876500337       &0.999999979353103      \\
$\frac{|c-1.00|}{1.00}$     &1.26E-07                 &4.61E-08              &\textbf{4.39E-09}       &1.23E-07       &1.33E-08      \\
$F_{Avg\pm Std}$  &1.07E-05$\pm$5.46E-06   &4.75E-07$\pm$2.74E-07   &\textbf{7.46E-08$\pm$3.29E-08}
       &1.89E-06$\pm$9.38E-07       &1.03E-07$\pm$6.12E-08  \\
\bottomrule
\end{tabular*}
\end{scriptsize}
\end{center}
\end{table}

The validation of the proposed methods in this paper is further proved by comparing CSML, CSP, CSC, and CSW with the standard CS algorithm for parameter identification. In the simulations, the maximum iteration number is set to 200 and the population size is set to 40. For the system to be identified, the step size is set to 0.005, and the number of samples set to 200. In addition, it is worth mentioning that the same computation effort is used in implementation for all the compared algorithms to make a fair comparison. Table~\ref{tabAPP} lists the statistical results of the average identified values, the corresponding relative error values, and the objective function values for system (\ref{FOFS}). From Table~\ref{tabAPP}, it can be clearly observed that all the four proposed randomness-enhanced CS algorithms outperform CS according to the average objective function values, and they are able to generate estimated values with much higher accuracy than CS. Besides, it can be seen that CSP surpasses CS, CSML, CSW, and CSC in obtaining the best average identified values, the corresponding relative error values, and the objective function values.

\begin{figure}[htbp]
\centering
\subfigure[]{
%\subfigure[pic1.]{
\begin{minipage}[t]{0.45\linewidth}
\centering
\includegraphics[width=3.in]{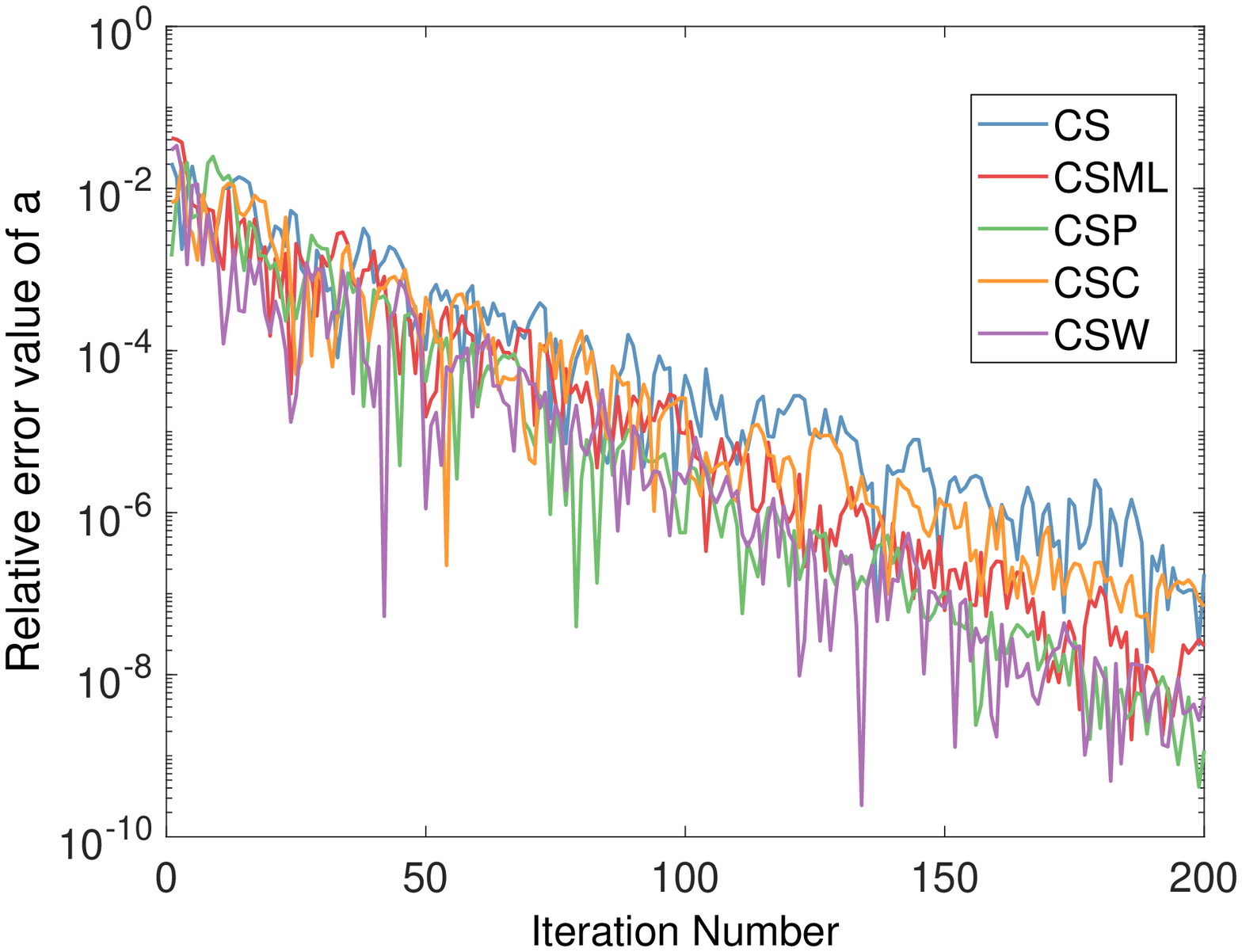}
\label{IDa}
\end{minipage}%
}%
\subfigure[]{
%\subfigure[pic2.]{
\begin{minipage}[t]{0.45\linewidth}
\centering
\includegraphics[width=3.in]{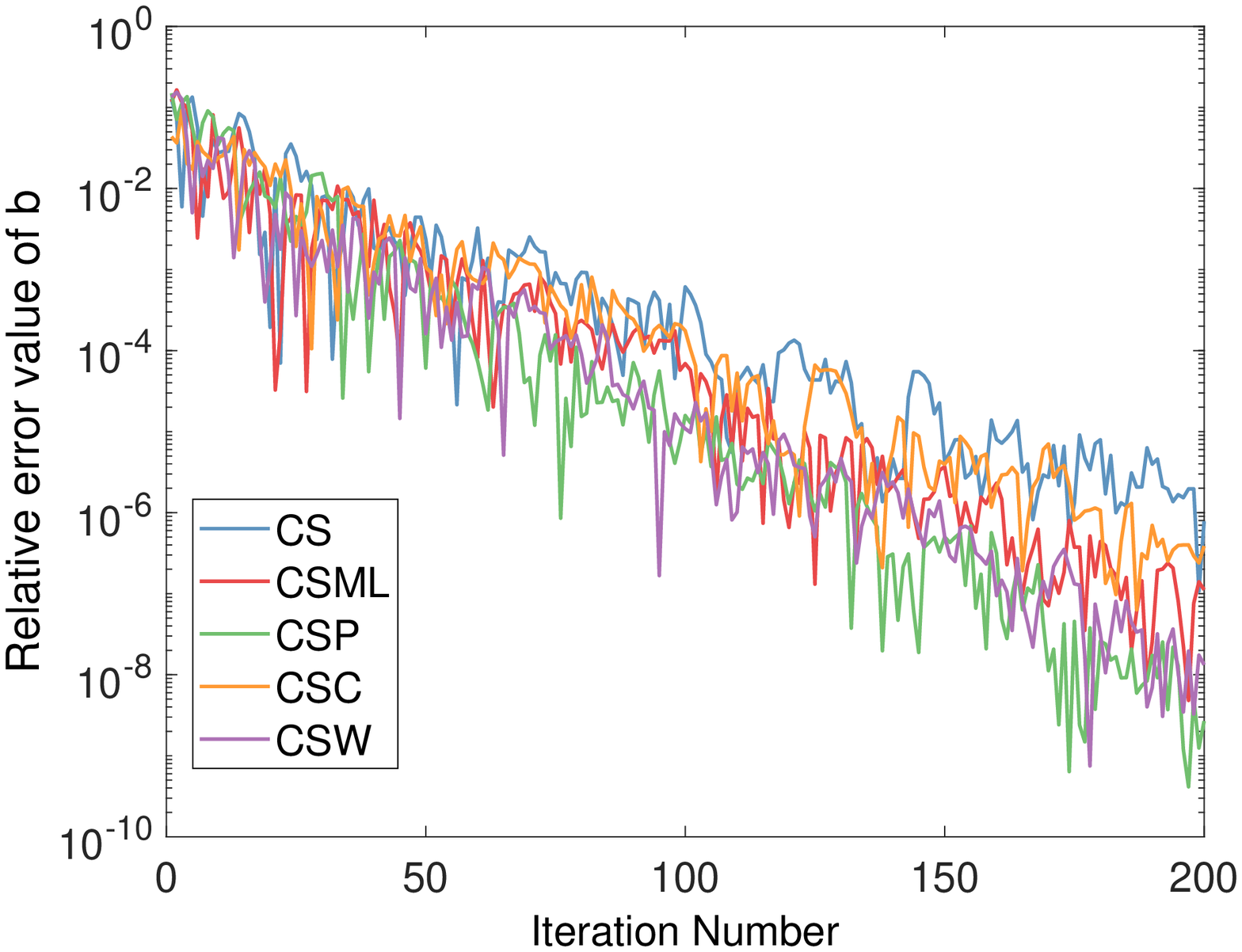}
\label{IDb}
\end{minipage}%
}%
\\
\subfigure[]{
\begin{minipage}[t]{0.45\linewidth}
\centering
\includegraphics[width=3.in]{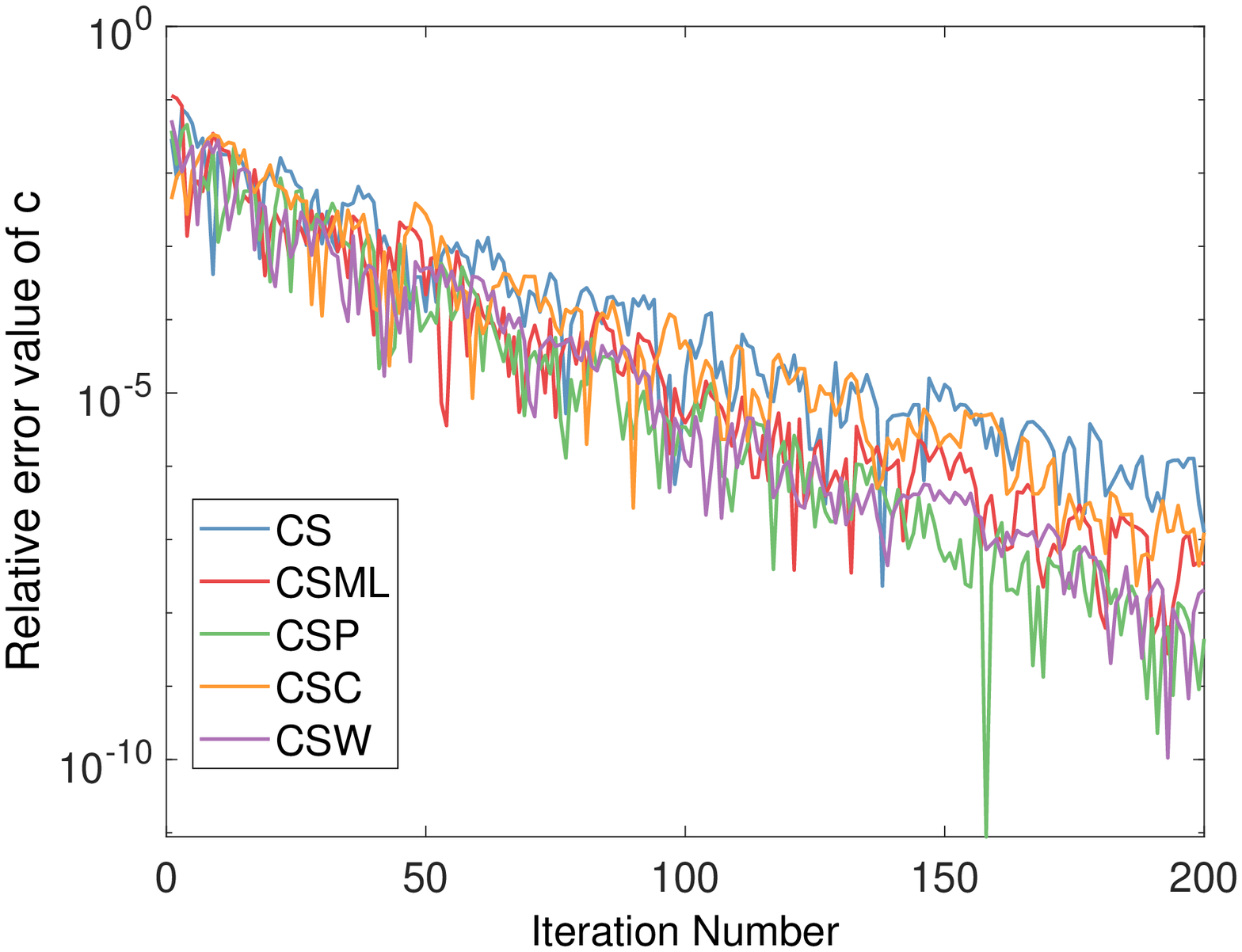}
\label{IDc}
\end{minipage}%
}%
\subfigure[]{
%\subfigure[pic2.]{
\begin{minipage}[t]{0.45\linewidth}
\centering
\includegraphics[width=3.in]{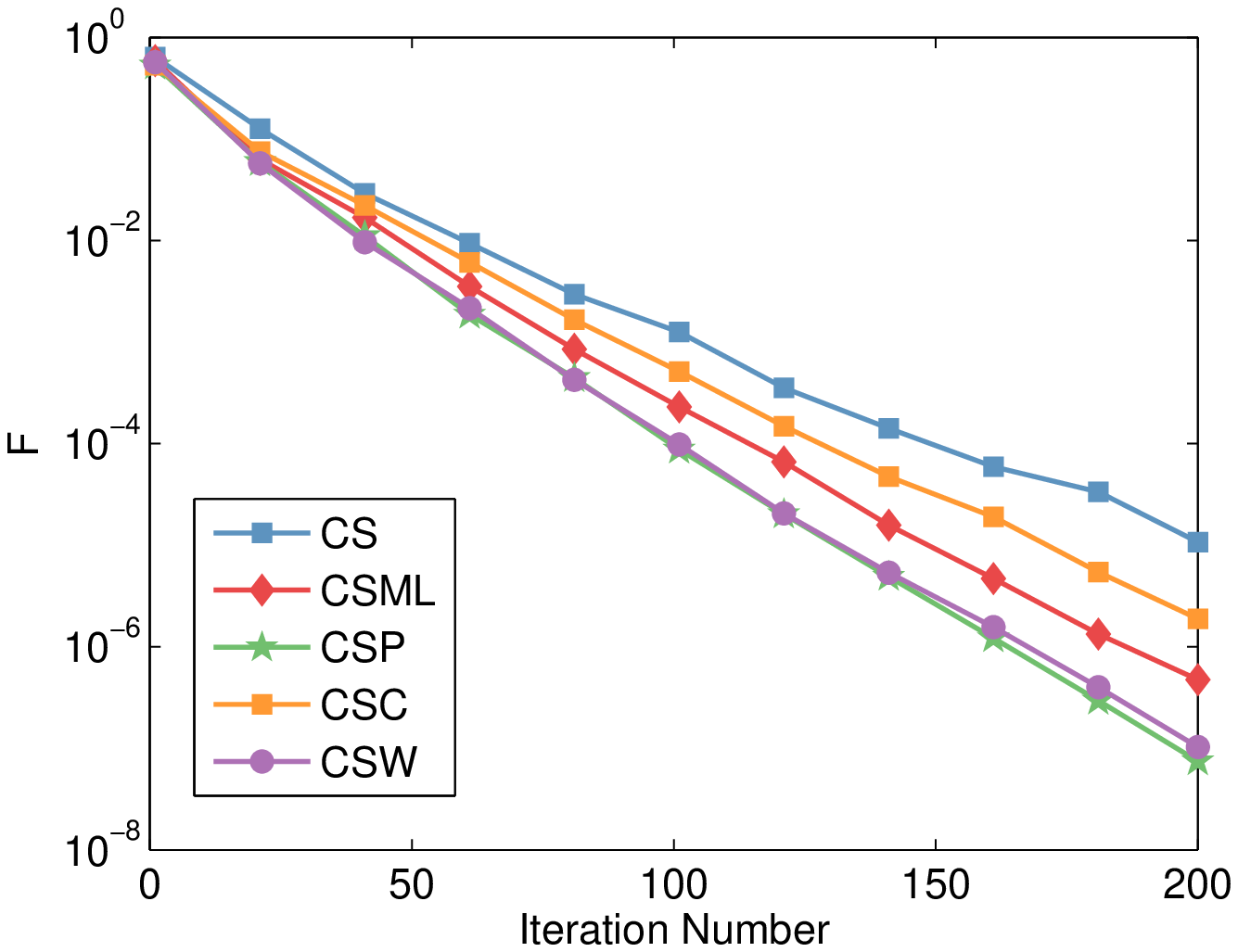}
\label{IDF}
\end{minipage}%
}%
\caption{The convergence curves of the relative error values and objective function values for system (\ref{FOFS}).}
\label{figcompare2}
\end{figure}

Moreover, Figure~\ref{figcompare2} shows the convergence curves of the relative error values of the estimated parameters and objective function values for the corresponding system via CSML, CSP, CSC, CSW, and CS. From Figures~\ref{IDa}~\ref{IDb}~\ref{IDc}, the relative error values of the estimated values generated by the randomness-enhanced CS algorithms converge to zero more quickly than the original CS. This indicates that CS algorithms with the four different heavy-tailed probability distributions are able to obtain more accurate values of the estimated parameters. In terms of Figure~\ref{IDF}, the objective function values of CSML, CSP, CSC, CSW also decline faster than CS, and among which CSP performs the best. It is noteworthy that CSW has a similar convergence curve of objective function values with CSP, and can converge to the nearby area of CSP. Therefore, CSW can still be considered as an efficient tool for solving optimization problems.

According to the foregoing discussion, it can be summarized that the randomness-enhanced CS algorithms are able to exactly identify the unknown specific parameters of the fractional-order system (\ref{FOFS}) with better effectiveness and robustness, and CSP together with CSW may be treated as a useful tool for handling the problem of parameter identification.

%%%%%%%%%%%%%%%%%%%%%%%%%%%%%%%%%%%%%%%%%%
\section{Conclusions}\label{CF}
The purpose of this paper is to discuss the optimal randomness in swarm-based search algorithms. In the study, CS is taken as a representative method of swarm-based optimization algorithms, and the results can be generalized to other swarm-based search algorithms. The impact of different heavy-tailed distributions on the performance of CS is investigated. By replacing L\'{e}vy flights with steps generated from other heavy-tailed distributions in CS, four different randomness-enhanced CS algorithms (namely CSML, CSP, CSC, and CSW) are presented by applying Mittag-Leffler distribution, Pareto distribution, Cauchy distribution and Weibull distribution, in order to improve the optimization performance of CS. The improvement in effectiveness and efficiency is validated through dedicated experiments. The experimental results indicate that all four proposed randomness-enhanced CS algorithms show a significant improvement in effectiveness and efficiency over the standard CS algorithm. Furthermore, the randomness-enhanced CS algorithms are successfully applied to system identification. In summary, CS with different heavy-tailed probability distributions can be regarded as an efficient and promising tool for solving the real-world complex optimization problems besides the benchmark problems.
% Since the current work is based on experimental verifications, we plan to give a theoretical analysis of global convergence of randomness-enhanced CS algorithms in the future, and make further improvements to randomness-enhanced CS algorithms to make it more comparable to other state-of-the-art algorithms.

Future promising topics can be directed to 1) theoretically analyze the global convergence of randomness-enhanced CS algorithms; 2) do a similar analyses to other swarm-based search algorithms for the optimal randomness; 3) since the search range is always finite for swarm-based search algorithms, it is necessary to study the optimal randomness in a finite range.

%%%%%%%%%%%%%%%%%%%%%%%%%%%%%%%%%%%%%%%%%%

%%%%%%%%%%%%%%%%%%%%%%%%%%%%%%%%%%%%%%%%%%

\vspace{6pt}


\begin{thebibliography}{999}
% Reference 1
\bibitem{yang2010nature}
Xin-She Yang. {\em Nature-inspired metaheuristic algorithms}. Luniver press, 2010.

\bibitem{anandakumar2018bio}
H Anandakumar and K Umamaheswari. A bio-inspired swarm intelligence technique for social aware cognitive radio handovers. \emph{Computers \& Electrical Engineering}, 71: 925-937, 2018.

\bibitem{brezovcnik2018swarm}
Lucija Brezo{\v{c}}nik, Iztok Fister, and Vili Podgorelec. Swarm intelligence algorithms for feature selection: a review. \emph{Applied Sciences}, 8(9): 1521, 2018.

\bibitem{zhao2019research}
Xuejing Zhao, Chen Wang, Jinxia Su, and Jianzhou Wang. Research and application based on the swarm intelligence algorithm and artificial intelligence for wind farm decision system. \emph{Renewable Energy}, 134: 681-697, 2019.


\bibitem{dulebenets2017novel}
Maxim A Dulebenets. A novel memetic algorithm with a deterministic parameter control for efficient berth scheduling at marine container terminals. \emph{Maritime Business Review}, 2(4): 302-330, 2017.


\bibitem{karaboga2007powerful}
Dervis Karaboga and Bahriye Basturk. A powerful and efficient algorithm for numerical function optimization: artificial bee colony (abc) algorithm. \emph{Journal of global optimization}, 39(3): 459-471, 2007.


\bibitem{yang2009cuckoo}
Xin-She Yang and Suash Deb. Cuckoo search via L\'{e}vy flights. In \emph{Nature \& Biologically Inspired Computing}, pages 210-214. IEEE, 2009.


\bibitem{yang2009firefly}
Xin-She Yang. Firefly algorithms for multimodal optimization. In \emph{International symposium on stochastic algorithms}, pages 169-178. Springer, 2009.


\bibitem{kennedy1995particle}
J Kennedy and R Eberhart. Particle swarm optimization (pso). In \emph{Proc. IEEE International Conference on Neural Networks, Perth, Australia}, pages 1942-1948, 1995.


\bibitem{zheng2012novel}
Hongqing Zheng and Yongquan Zhou. A novel cuckoo search optimization algorithm based on gauss distribution. \emph{Journal of Computational Information Systems}, 8(10): 4193-4200, 2012.


\bibitem{wang2016nearest}
Lijin Wang, Yiwen Zhong, and Yilong Yin. Nearest neighbour cuckoo search algorithm with probabilistic mutation. \emph{Applied Soft Computing}, 49: 498-509, 2016.


\bibitem{rakhshani2017snap}
Hojjat Rakhshani and Amin Rahati. Snap-drift cuckoo search: A novel cuckoo search optimization algorithm. \emph{Applied Soft Computing}, 52: 771-794, 2017.


\bibitem{cui2017novel}
Zhihua Cui, Bin Sun, Gaige Wang, Yu Xue, and Jinjun Chen. A novel oriented cuckoo search algorithm to improve dv-hop performance for cyber–physical systems. \emph{Journal of Parallel and Distributed Computing}, 103: 42-52, 2017.


\bibitem{salgotra2018new}
Rohit Salgotra, Urvinder Singh, and Sriparna Saha. New cuckoo search algorithms with enhanced exploration and exploitation properties. \emph{Expert Systems with Applications}, 95: 384-420, 2018.


\bibitem{richer2006levy}
Toby J Richer and Tim M Blackwell. The L\'{e}vy particle swarm. In \emph{2006 IEEE International Conference on Evolutionary Computation}, pages 808-815. IEEE, 2006.

\bibitem{pavlyukevich2007levy}
Ilya Pavlyukevich. L\'{e}vy flights, non-local search and simulated annealing. \emph{Journal of Computational Physics}, 226(2): 1830-1844, 2007.

\bibitem{yang2014nature}
Xin-She Yang. \emph{Nature-inspired optimization algorithms}. Elsevier, 2014.

\bibitem{yang2010engineering}
Xin-She Yang and Suash Deb. Engineering optimisation by cuckoo search. \emph{International Journal of Mathematical Modelling and Numerical Optimisation}, 1(4): 330-343, 2010.

\bibitem{foss2011introduction}
Sergey Foss, Dmitry Korshunov, Stan Zachary, et al. \emph{An introduction to heavy-tailed and subexponential distributions}, volume 6. Springer, 2011.


\bibitem{kozubowski1999univariate}
Tomasz J Kozubowski and Svetlozar T Rachev. Univariate geometric stable laws. \emph{Journal of Computational Analysis and Applications}, 1(2): 177-217, 1999.


\bibitem{noman2008accelerating}
Nasimul Noman and Hitoshi Iba. Accelerating differential evolution using an adaptive local search. \emph{IEEE Transactions on evolutionary Computation}, 12(1): 107-125, 2008.

\bibitem{suganthan2005problem}
Ponnuthurai N Suganthan, Nikolaus Hansen, Jing J Liang, Kalyanmoy Deb, Ying-Ping Chen, Anne Auger, and Santosh Tiwari. Problem definitions and evaluation criteria for the cec 2005 special session on realparameter optimization. \emph{KanGAL report}, 2005005: 2005, 2005.

\bibitem{storn1997differential}
Rainer Storn and Kenneth Price. Differential evolution-a simple and efficient heuristic for global optimization over continuous spaces. \emph{Journal of global optimization}, 11(4): 341-359, 1997.

\bibitem{yang2012flower}
Xin-She Yang. Flower pollination algorithm for global optimization. In \emph{International conference on unconventional computing and natural computation}, pages 240-249. Springer, 2012.

\bibitem{gao2014inversion}
Fei Gao, Feng-xia Fei, Xue-jing Lee, Heng-qing Tong, Yan-fang Deng, and Hua-ling Zhao. Inversion mechanism with functional extrema model for identification incommensurate and hyper fractional chaos via differential evolution. \emph{Expert Systems with Applications}, 41(4): 1915-1927, 2014.


\bibitem{Chen2008Nonlinear}
Wei-Ching Chen. Nonlinear dynamics and chaos in a fractional-order financial system. \emph{Chaos, Solitons \& Fractals}, 36(5): 1305-1314, 2008.


\end{thebibliography}
\end{document}